\documentclass[journal]{IEEEtran}
\ifCLASSINFOpdf
\else
\fi
\usepackage[ruled,vlined]{algorithm2e}

\usepackage{multirow}
\usepackage{booktabs}
\usepackage{mathtools}
\usepackage{graphicx}
\usepackage{amsmath}
\usepackage{cite}
\usepackage{bm}
\usepackage{subfigure}
\usepackage{xcolor}

\begin{document}
%
\title{Mixed Precision Low-bit Quantization of Neural Network Language Models for Speech Recognition}
%
%
%

\author{Junhao~Xu, Jianwei~Yu, Shoukang~Hu,
            Xunying~Liu,~\IEEEmembership{Member,~IEEE},
            Helen~Meng,~\IEEEmembership{Fellow,~IEEE}
\thanks{Junhao Xu, Jianwei Yu, Shoukang Hu, Xunying Liu, and Helen Meng are with the Chinese University of Hong Kong, Hong Kong 999077, China (e-mail: jhxu@se.cuhk.edu.hk; jwyu@se.cuhk.edu.hk; skhu@se.cuhk.edu.hk; xyliu@se.cuhk.edu.hk; hmmeng@se.cuhk.edu.hk. Corresponding author: Xunying Liu.}
}

%
%

\markboth{Journal of \LaTeX\ Class Files,~Vol.~14, No.~8, August~2015}%
{Shell \MakeLowercase{\textit{et al.}}: Bare Demo of IEEEtran.cls for IEEE Journals}
%



\maketitle

\begin{abstract}
  State-of-the-art language models (LMs) represented by long-short term memory recurrent neural networks (LSTM-RNNs) and Transformers are becoming increasingly complex and expensive for practical applications. Low-bit neural network quantization provides a powerful solution to dramatically reduce their model size. Current quantization methods are based on uniform precision and fail to account for the varying performance sensitivity at different parts of LMs to quantization errors. To this end, novel mixed precision neural network LM quantization methods are proposed in this paper. The optimal local precision choices for LSTM-RNN and Transformer based neural LMs are automatically learned using three techniques. The first two approaches are based on quantization sensitivity metrics in the form of either the KL-divergence measured between full precision and quantized LMs, or Hessian trace weighted quantization perturbation that can be approximated efficiently using matrix free techniques. The third approach is based on mixed precision neural architecture search. In order to overcome the difficulty in using gradient descent methods to directly estimate discrete quantized weights, alternating direction methods of multipliers (ADMM) are used to efficiently train quantized LMs. Experiments were conducted on state-of-the-art LF-MMI CNN-TDNN systems featuring speed perturbation, i-Vector and learning hidden unit contribution (LHUC) based speaker adaptation on two tasks: Switchboard telephone speech and AMI meeting transcription. The proposed mixed precision quantization techniques achieved “lossless” quantization on both tasks, by producing model size compression ratios of up to approximately 16 times over the full precision LSTM and Transformer baseline LMs, while incurring no statistically significant word error rate increase.
\end{abstract}

\begin{IEEEkeywords}
  Language models, Speech recognition, LSTM-RNN, Transformer, Low-bit Quantization, ADMM 
\end{IEEEkeywords}

%
\IEEEpeerreviewmaketitle

\section{Introduction}

\IEEEPARstart{L}{anguage} models (LMs) are important components in automatic speech recognition (ASR) systems and many other applications. A key part of the statistical language modelling problem is to derive the suitable representation of long-range history contexts. Directly modelling long-span word histories using conventional back-off $n$-gram models~\cite{Slave-katz-1987} generally leads to a severe data sparsity issue~\cite{Stanley-1999}. To this end, over the past few decades there have been significant efforts of developing artificial neural network based language modelling techniques in the speech technology community~\cite{Bengio-2003,Schwenk-2007,Mikolov-2010,Ebru-Tara,Yvon-2013,Sundermeyer-2015,XChen-2016,XChen-2019,Irie-2019,Keli-2019,Beck-ralf,Pau-jorge}. Neural network language models (NNLMs) representing longer span history contexts in a continuous and lower dimensional vector space, are used to improve the generalization performance.

With the rapid progress of deep neural network (DNN) based ASR technologies in recent decades, the underlying network architectures of NNLMs have evolved from feedforward structures~\cite{Bengio-2003,Schwenk-2007,Mikolov-2010,Ebru-Tara,Yvon-2013} to more advanced variants represented by long-short term memory recurrent neural networks (LSTM-RNNs)~\cite{LSTM-1997,Sundermeyer-2015,XChen-2016,XChen-2019} and more recently neural Transformers~\cite{Vaswani-2017,Irie-2019,Keli-2019,Beck-ralf,Pau-jorge} that are designed for modelling longer range contexts. In particular, Transformer based language models in recent years have defined state-of-the-art performance across a range of ASR task domains~\cite{Irie-2019,Keli-2019,Beck-ralf,Pau-jorge,Jun-2021}. These models~\cite{Irie-2019,Keli-2019,Beck-ralf,Jun-2021} are often constructed using a deep stacking of multiple self-attention based neural building blocks~\cite{Cheng-2016,Zlin-2017,Parikh-2016}, each of which also includes residual connections~\cite{Khe-2017}
 and layer normalization modules~\cite{JLBa-2016}. Additional positional encoding 
 layers~\cite{Vaswani-2017,Gehring-2017} are employed to augment the self-attention structures with word sequence order information. Performance improvements over conventional LSTM-RNN language models have been widely 
 reported~\cite{Irie-2019,Zeyer-2019}. 

However, the increasingly deeper and more complex architecture designs featuring in LSTM-RNN and Transformer models present many challenges for current ASR technologies. These not only lead to a large increase in their overall memory footprint and computational cost when operating on the cloud, but also creates difficulty when deployed on edge devices to enhance privacy and reduce latency. In a wider context within the speech technology community, such dramatically increasing demand for computational resources is consistent with the recent trend of moving towards a data and computational intensive all neural end-to-end (E2E) modelling paradigm represented by, for example, 
Transformers~\cite{Wang-2020,Zhang-2020,Karita-2019}, RNN transducers (RNN-T)~\cite{Graves-2012}, and listen, attend and spell (LAS)~\cite{Chan-2016}. State-of-the-art ASR systems featuring these end-to-end approaches often contain a very large number of parameters, for example, up to 280 million~\cite{Tuske-2020}. Hence, there is a pressing need of developing ultra-compact, low footprint language modelling methods, and ASR technologies in general, to facilitate more aggressive reduction in memory footprint, model training and evaluation time while maintaining competitive accuracy performance. 

To this end, significant efforts have been made in both the machine learning and speech technology communities to develop DNN compression techniques~\cite{Soudry-2014,Courbariaux-2015,Hubara-2017,Rastegari-2016,ZDong-2019,KWang-2019,ZDong-2019-2,SHan-2015,SHan-2016,HMao-2017,BLiu-2015,WWen-2016,Szegedy-2016,Szegedy-2017,Hinton-2014,Romero-2015,Chebotar-2016,Jaderberg-2014,Lebedev-2015,CTai-2016,Sindhwanietal-2015,Iandola-SqueezeNet,XZhang-2018,NMa-2018,Sainath-2013,YWang-2015,XLiu-2018,KYu-2020,YHe-2019,Dudziak-2019,Sim-2019,YQian-2019,Stanley-2002,Kandasamy-2018,Zoph-2017,Baker-2017,HCai-2018,ZZhong-2018,HLiu-2018,SXie2019,HCai-2019,HCai-2020,Boyd-2011,CLeng-2018}. 
Pruning based methods exploiting the structural and parameter sparsity were used to reduce DNN model size in~\cite{SHan-2015,SHan-2016,HMao-2017,BLiu-2015,WWen-2016,Szegedy-2016,Szegedy-2017}. They are particularly useful for models containing large fully connected layers such as the ResNet systems~\cite{Szegedy-2016,Szegedy-2017}. Knowledge distillation and teacher-student learning~\cite{Hinton-2014,Romero-2015,Chebotar-2016} based approaches extract information from a pre-trained, larger model into a smaller one. Low rank matrix 
factorization~\cite{Jaderberg-2014,Lebedev-2015,CTai-2016,Sindhwanietal-2015,Sainath-2013}, and neural architecture search (NAS) based 
methods~\cite{Stanley-2002,Kandasamy-2018,Zoph-2017,Baker-2017,HCai-2018,ZZhong-2018,HLiu-2018,SXie2019,HCai-2019,HCai-2020} have also been proposed. 

Another powerful family of techniques recently drawing increasing interest across the machine learning, computer vision and speech technology communities to solve this problem is to use low-bit DNN quantization techniques~\cite{Soudry-2014,Courbariaux-2015,Hubara-2017,Rastegari-2016,ZDong-2019,KWang-2019,ZDong-2019-2,Iandola-SqueezeNet,XLiu-2018,KYu-2020,YQian-2019,CLeng-2018,RMa-2019}. By replacing floating point based DNN parameters with low precision values, for example, binary numbers, model sizes can be dramatically reduced without changing the DNN architecture~\cite{XLiu-2018,Courbariaux-2015,Boyd-2011}. Further DNN size reduction can be obtained when low-precision quantization is used in combination with neural architecture search (NAS) techniques, for example, in the SqueezeNet system designed for computer vision tasks~\cite{Iandola-SqueezeNet}. In contrast to the extensive prior research works on low-bit quantization methods primarily targeting computer vision 
tasks~\cite{Soudry-2014,Courbariaux-2015,Hubara-2017,Rastegari-2016,ZDong-2019,KWang-2019,ZDong-2019-2,Iandola-SqueezeNet}, only limited previous research in this direction has been conducted in the context of language modelling~\cite{XLiu-2018,KYu-2020} and ASR systems~\cite{YWang-2015,YHe-2019}. 

Two issues are associated with current low-bit DNN quantization methods. First, these approaches are predominantly based on uniform precision, where an identical bit-width is applied to all weight parameters for quantization~\cite{Boo-2020}. Within such framework, the varying local performance sensitivity exhibited at different parts of the underlying DNN system to quantization errors is not taken into account. In practice, this often leads to large performance degradation against full precision models. Second, when DNN weights are restricted to discrete values, the conventional back-propagation (BP) algorithm based on gradient descent methods cannot be directly applied to estimate the quantized model parameters. Existing approaches for training quantized DNNs often use a modified BP algorithm~\cite{Soudry-2014,Courbariaux-2015}. In this approach, low precision quantized parameters are first used in a forward pass to compute the error loss, before full precision parameters are then used in a backward pass to propagate the gradients for subsequent model update. However, the inconsistency between quantized, discrete weights and the SGD algorithm assuming continuous and differentiable error cost functions leads to not only very slow convergence in training, but also performance degradation against full precision models. 

In order to address the first issue discussed above regarding performance sensitivity, and motivated by the recent development of mixed precision DNN acceleration hardware that allows multiple locally selected precision settings to be used~\cite{ZDong-2019-2}, novel mixed precision DNN quantization approaches are proposed in this paper by utilizing locally variable bit-widths at different layer components of LSTM-RNN and Transformer LMs. The optimal local precision settings are automatically learned using three techniques. The first two approaches are based on quantization sensitivity metrics in the form of either Hessian trace weighted quantization perturbation that can be approximated efficiently via matrix free techniques, or the KL-divergence measured between full precision and quantized language models. The third approach is based on mixed precision neural architecture search. 

In order to address the second issue over the difficulty in using gradient descent methods to directly estimate NNLMs of discrete weights, the general problem of estimating quantized DNN model parameters is reformulated as an optimization task. For any form of quantized NNLMs using uniform or mixed precision settings, alternating direction methods of multipliers (ADMM)~\cite{Boyd-2011,Jun-2021,Jun-2020} are proposed to efficiently train their discrete parameters. Two sets of model parameters respectively associated with a full precision neural network LM, and the corresponding optimal quantized model with a particular precision setting, are iteratively learned via a decomposed dual ascent scheme in an alternating fashion. This novel quantized NNLM estimation algorithm draws strength from both the decomposability of classic dual ascent schemes and the stable convergence of multiplier methods. 

Experiments were conducted on state-of-the-art LF-MMI CNN-TDNN and TDNN systems~\cite{Povey2016} featuring speed perturbation, i-Vector~\cite{Dehak2011} and learning hidden unit contribution (LHUC) based speaker adaptation~\cite{swietojanski2014learning} on two tasks, Switchboard telephone speech~\cite{Godfrey-SWBD} and AMI meeting transcription~\cite{Hain-AMI}. Experimental results suggest the proposed mixed precision LM quantization techniques achieved model size compression ratios of about 16 times over the full precision Transformer LM baselines with no statistically significant recognition performance degradation. 

The main contributions of this paper are summarized as follows. 

1) To the best of our knowledge, this paper presents the first work in the speech technology community to apply mixed precision DNN quantization techniques to both LSTM-RNN and Transformer based NNLMs. In contrast, prior researches within the speech community in this direction largely focused on uniform precision based quantization of convolutional neural networks (CNNs) acoustic models~\cite{YQian-2019} and LSTM-RNN language models~\cite{KYu-2020,XLiu-2018,RMa-2019}. 

2) To the best of our knowledge, this paper is the first work to introduce ADMM based neural network quantization techniques for speech recognition tasks. In contrast, prior researches with the speech technology community on low-bit quantization of CNNs~\cite{YQian-2019} and LSTM-RNN LMs~\cite{KYu-2020,XLiu-2018,RMa-2019} used the modified BP algorithm~\cite{Soudry-2014,Courbariaux-2015} while the inconsistency between discrete, quantized parameters and gradient based SGD update remains unaddressed. 

3) Transformer LM model size compression ratios of up to approximately 16 times over the full precision Transformer LM baselines with no statistically significant recognition performance degradation were obtained using an average 2-bit mixed precision configuration. To the best of our knowledge, this is the best low-bit Transformer language model compression ratio published so far in the speech technology community while incurring no recognition accuracy loss. 

The rest of the paper is organized as follows. LSTM-RNN and Transformer based NNLMs are reviewed in section II. A general neural network quantization scheme based on uniform or locally varying mixed quantization precisions are presented in section III. ADMM based training of quantized NNLMs are presented in Section IV. Section V presents three novel mixed precision quantization methods. Experiments and results are shown in Section VI. Finally, conclusions and possible future work are discussed in section VII. 

\section{NN Language Models}
This section reviews two types of neural network language models that are widely used in state-of-the-art speech recognition systems: long-short term memory recurrent neural network (LSTM-RNN) and Transformer based language models.

\vspace{-1em}
\subsection{Recurrent Neural Network LMs}
The form of LSTM-RNN language models considered in this paper computes the word probability $\mathbf{w}_t$ given the preceding history context of $t-1$ words $\mathbf{w}_1,...,\mathbf{w}_{t-1}$ as
 \begin{align}
  \vspace{-2cm} 
      P(\mathbf{w}_{t}|\mathbf{w}^{t-1}_{1})\approx P\left(\mathbf{w}_{t}|\mathbf{w}_{t-1}, \mathbf{h}_{t-1}\right),
  \vspace{-1cm} 
  \end{align}
  
 \noindent where $\mathbf{h}_{t-1}$ is the D-dimensional vector \textit{hidden state} encoding part of the history information 
 $(\mathbf{w}_{1},...,\mathbf{w}_{t-2})$ up to word $\mathbf{w}_{t-2}$, where D is the number of RNNLM hidden layer nodes. The most recent word history $\mathbf{w}_{t-1}$ is represented by a $N$-dimensional one-hot vector $\mathbf{\tilde{w}}_{t}$, where $N$ is the vocabulary size. This one-hot word input vector is first projected into a $M$-dimensional ($M\ll N$) linear vector embedding as
  \begin{align}
    \vspace{-2cm} 
    \mathbf{x}_{t}=\mathbf{\Theta}_U\mathbf{\tilde{w}}_{t}^\top,
    \vspace{-1cm} 
    \end{align}

\noindent where $\mathbf{\Theta}_U$ is a projection matrix to be learned, before being further fed into the hidden layers where non-linear transformations are applied. The hidden state $\mathbf{h}_t$ is calculated within the LSTM cells ~\cite{Hochreiter-LSTM} where the previous hidden state $\mathbf{h}_{t-1}$ and the current word input embedding $\mathbf{\tilde{w}}_{t}$ is combined. Each LSTM memory cell consists of a set of Sigmoidal gating activations, the input gate $\mathbf{i}_t$,  forget gate $\mathbf{f}_t$, cell gate $\mathbf{\tilde c}_t$ and output gate $\mathbf{o}_t$. These are used to control the information flow within the cells in order to trap longer range history contexts and address the vanishing gradient issue. The respective outputs from these gates are given by
\begin{align}
\vspace{-1cm} 
    \label{eq:f-gate}
    \mathbf{f}_t=&\, \boldsymbol{\sigma}\left(\boldsymbol{\Theta}_\text{f}\left[\mathbf{x}_{t-1}, \mathbf{h}_{t-1}, 1\right]^\top\right)\\
    \label{eq:i-gate}
    \mathbf{i}_t=&\, \boldsymbol{\sigma}\left(\boldsymbol{\Theta}_\text{i}\left[\mathbf{x}_{t-1}, \mathbf{h}_{t-1}, 1\right]^\top\right)\\
    \label{eq:c-gate}
    \mathbf{\tilde{c}}_t=&\, \text{tanh}\left(\boldsymbol{\Theta}_\text{c}\left[\mathbf{x}_{t-1}, \mathbf{h}_{t-1}, 1\right]^\top\right)\\
    \label{eq:o-gate}
    \mathbf{o}_t=&\, \boldsymbol{\sigma}\left(\boldsymbol{\Theta}_\text{o}\left[\mathbf{x}_{t-1}, \mathbf{h}_{t-1}, 1\right]^\top\right)
\vspace{-1cm} 
\end{align}
\noindent where $\text{tanh}(\mathbf{u})=[\text{tanh}(\mathbf{u}_1), ..., \text{tanh}(\mathbf{u}_D)]$ for any $\mathbf{u}\in\mathbf{R}^D$.

The final hidden state is normally computed recursively in an auto-regressive fashion, for example, from left to right in case of standard uni-directional LSTM-RNN LMs modelling history contexts only~\cite{Mikolov-2010, Sundermeyer-2015,XChen-2019}. Using the above four gating functions outputs, the LSTM-RNN cell contents ${\bf c}_t$ and hidden state representation ${\bf h}_t$ are finally computed as
\begin{align}
\vspace{-2em}
    \label{eq:c_t}
    \mathbf{c}_t=&\,\mathbf{f}_t\otimes \mathbf{c}_{t-1} + \mathbf{i}_t\otimes \mathbf{\tilde{c}}_t\\
    \label{eq:h_t}
    \mathbf{h}_{t}=&\,\mathbf{o}_t\otimes\tanh(\mathbf{c}_t),
\vspace{-1cm} 
\end{align}
where $\otimes$ is the Hadamard product. 

\vspace{-1em}
\subsection{Transformer LMs}
The Transformer model architecture considered in this paper features a deep stacking of multiple Transformer decoder blocks. Unless otherwise stated, 6 Transformer decoder blocks are used in all the experiments of this paper. As shown in the top part of Figure 1, each Transformer decoder block consists of a multi-head self-attention~\cite{Cheng-2016,Zlin-2017,Parikh-2016} module and a feed forward module. Residual connections~\cite{Khe-2017} and layer normalization operations~\cite{JLBa-2016} are also inserted between these two modules. Let $\mathbf{x}_t^{l-1}$ denotes the output of the $(l-1)$-th Transformer block at input word position $t$. The multi-head self-attention module in the succeeding $l$-th block transforms $\mathbf{x}_t^{l-1}$ to $\mathbf{z}_t^l$ as follows:
\begin{align}
    {\bf q}_{t}^{l}, {\bf k}_{t}^{l}, {\bf v}_{t}^{l}&=\mathbf{\Theta}_Q^l{\bf x}_{t}^{l-1},\mathbf{\Theta}_K^l{\bf x}_{t}^{l-1},\mathbf{\Theta}_V^l{\bf x}_{t}^{l-1} \\
    {\bf y}_{t}^{l} &= \text{Attention}(\boldsymbol{k}_1^l,...,\boldsymbol{k}_t^l, \boldsymbol{v}_1^l, ...,\boldsymbol{v}_t^l, {\bf q}_{t}^{l}) \\
    {\bf z}_{t}^{l}&=\mathbf{\Theta}_h^{l}{\bf y}_{t}^{l} + {\bf x}_{t}^{l-1} \\
    {\bf o}_{t}^{l} &= \text{LayerNorm}({\bf z}_{t}^{l})
\end{align}
\noindent where $\mathbf{\Theta}_Q^l, \mathbf{\Theta}_K^l, \mathbf{\Theta}_V^l$ denote the learnable query, key, value projection matrices which map $\mathbf{x}_t^{l-1}$ into the corresponding vector representations of query ${\bf q}_{t}^{l}$, key ${\bf k}_{t}^{l}$ and value ${\bf v}_{t}^{l}$ respectively. ${\bf z}_{t}^{l}$ is the sequence of of cached key-value vector pairs up to time $t$, which only contains the history context information and can prevent the model from using any future context. ${\text{Attention}(\cdot)}$ denotes the scaled multi-head dot product self-attention~\cite{Parikh-2016}. ${\text{LayerNorm}(\cdot)}$ represents the layer normalization operation~\cite{JLBa-2016}. ${\bf \Theta}_{h}^{l}$ denotes the learnable projection matrix applied to the outputs of the Attention operation prior to layer normalization. The normalized output ${\bf o}_{t}^{l}$ is then fed into the feed forward module: 
\begin{align}
    {\bf s}_{t}^{l} &= \mathbf{\Theta}_2^{l}\text{GELU}(\mathbf{\Theta}_1^l{\bf o}_{t}^{l}) + {\bf o}_{t}^{l} \\
    {\bf x}_{t}^{l} &= \text{LayerNorm}({\bf s}_{t}^{l})
\end{align}
\noindent where $\mathbf{\Theta}_1^{l}$ and  $\mathbf{\Theta}_2^{l}$ are the weight matrices that are applied to the normalized output ${\bf o}_{t}^{l}$ before Gaussian error linear unit (GELU)~\cite{Hendrycks-GELU} activation functions and after. For simplicity the bias vectors optionally used in the feed forward modules are omitted in the above Equation (13). In addition, positional embedding layers are also used in all the transformer LMs considered in this paper. 

\section{Neural Network Quantization}
The standard n-bit quantization problem for deep neural networks (DNNs) considers the task of for any full precision weight parameter, $\Theta$, finding its closest discrete approximation from the following quantization table with a global scaling factor $\alpha$ and  quantization parameter $V$, 
\begin{equation}
\label{equ:simQuan}
q = \alpha V \in \{0, \pm\alpha, \dots, \pm\alpha\cdot (2^{n-1}-1)\}
\end{equation}
\noindent as the one that incurs the minimum quantization error
\begin{equation}
  \begin{split}
  \label{eq:quanMap}
  f(\Theta) = \arg\min \limits_{q} |\Theta - q|
\vspace{-1cm} 
\end{split}
\end{equation}

Further simplification to the above quantization table of Equation (\ref{equ:simQuan}) leads to extremely low bit quantization based on, for example, binary values $\{-1, 1\}$~\cite{CLeng-2018,Rastegari-2016}, or tertiary values $\{-1, 0, 1\}$~\cite{LiF-2016}.

It is assumed in the above standard quantization process that a global quantization table is applied to all weight parameters. In order to account for the fine-grained local quantization sensitivity, the following more general form of quantization is considered for a particular model parameter  ${\Theta^{(l)}}$ within any of the $l$-th weight cluster, for example, all parameters of the same LSTM or Transformer LM layer,
 \begin{equation}
  \begin{split}
  \label{eq:quanMap}
  f(\Theta^{(l)}) = \arg\min \limits_{Q^{(l)}} |\Theta^{(l)} - Q^{(l)}|
\vspace{-1cm} 
\end{split}
\end{equation}
can be used. 
The locally shared $l$-th quantization table is
\begin{equation}
\label{equ:31}
  {Q}^{(l)} = \alpha^{(l)}V^{(l)} \in\{0, \pm\alpha^{(l)},\dots, \pm\alpha^{(l)} (2^{n_l-1}-1)\}
\end{equation}
\noindent where the full precision scaling factor $\alpha^{(l)}$ is used to adjust the dynamic range of all the quantized weights in the $l$-th cluster. It is shared locally within each individual DNN parameter clusters. The locally variable quantization bit length $n_{l}$ can be set to be $1, 2, 4, 8$ etc. depending on the optimal precision settings to be used. A special case, when the local quantization table in Equation (18) is shared across all the layers in the system, this leads to the traditional uniform precision quantization. The above tying of quantization tables may be flexibly performed at either layer, or node level, or in the extreme case at individual parameter level (this is equivalent to no quantization being applied). Intuitively, the longer the local quantization precision bit widths $\{n_{l}\}$ are used at each part of the underlying neural language model, a smaller compression ratio after quantization and reduced performance degradation is expected. 

\vspace{-0.5em}
\section{ADMM based Training of Quantized DNN}
One major challenge faced by both uniform and mixed precision quantization is that the gradient descent methods and back-propagation (BP) algorithm cannot be directly used when weights are quantized to discrete values. To this end, mixed precision BP was proposed in~\cite{Soudry-2014,Courbariaux-2015} where low precision binarized parameters were first used in the forward pass to compute the error loss before full precision parameters are used in the backward pass to propagate the gradients. However, there is an inconsistency between quantized, discrete weights and the assumption over continuous and differentiable error cost functions in SGD update. Hence, directly training quantized system using mixed precision BP leads to very slow convergence and the performance gap between full precision and quantized systems remains large. An alternative solution to this problem is to reformulate quantization as a constrained optimization problem solved by the alternating direction methods of multipliers (ADMM)~\cite{Boyd-2011}. 

The ADMM based optimization decomposes a dual ascent problem into alternating updates of two variables. In the context of the neural network LM quantization problem considered here, these correspond to the full precision model weights update and the discrete quantization tables estimation. The overall ADMM Lagrange function is given as 
\begin{equation}
\label{equ:AL}
  L = \mathcal{F}_{ce}(\boldsymbol{\Theta})+(\gamma \boldsymbol\lambda)^\top\cdot(\boldsymbol{\Theta}-f(\boldsymbol\Theta)) + \frac{\gamma}{2}||\boldsymbol{\Theta}-f(\boldsymbol\Theta)||_2^2
\end{equation}

\noindent where $\mathcal{F}_{ce}$ is the cross entropy (CE) loss, $\boldsymbol{\Theta}$ are the full precision model parameters. $f(\boldsymbol\Theta)$ represents the quantization of the parameters calculated from the quantization using Equation (17). $\gamma>0$ is the penalty parameter which is empirically set as $10^{-3}$ throughout this paper and $\boldsymbol\lambda$ denotes the Lagrangian multiplier. The standard Lagrangian term expressed in the form of a dot product between the multiplier variable $\boldsymbol\lambda$ and the quantization error, $\boldsymbol{\Theta}-f(\boldsymbol\Theta)$, is shown as the second term in Equation (19).  In order to further improve the robustness and convergence speed of the ADMM algorithm, an additional term related to the quantization error squared norm, shown as the third term in Equation (19), is also introduced to form an augmented Lagrangian~\cite{Boyd-2011}. Further rearranging Equation (19) leads to the following loss function. 
\begin{equation}
\label{equ:AL2}
  L = \mathcal{F}_{ce}(\boldsymbol{\Theta}) +  \frac{\gamma}{2}||\boldsymbol{\Theta}-f(\boldsymbol\Theta)+\boldsymbol{\lambda}||_2^2-\frac{\gamma}{2}||\boldsymbol{\lambda}||^2
\end{equation}
  
Given a particular uniform or mixed precision quantization configuration, the ADMM algorithm is iteratively performed to find the optimal scaling factors, $\{\alpha^{(l)}\}$, in the quantization table(s) of Equation (18). For simplicity, in the following detailed description of the algorithm, we assume a globally shared quantization table $\{0, \pm\alpha,\dots, \pm\alpha (2^{n-1}-1)\}$ is applied to all parameters and a single scaling factor $\alpha$ is to be learned. The following iterative update can be extended when multiple shared quantization tables of different bit-widths $n_l$ in Equation (18) are used. When the ADMM algorithm is performed at the $(k+1)^{th}$ iteration, the overall update can be split into three stages presented in the following subsections.

\vspace{-1em}
\subsection{Full precision model parameter update}
 \noindent The following equation is used to update the full precision weight parameters $\boldsymbol\Theta^{(k+1)}$.
\begin{equation}
  \boldsymbol{\Theta}^{(k+1)} = \arg\min \limits_{\boldsymbol{\Theta}} L(\boldsymbol{\Theta}, f(\boldsymbol\Theta^{(k)}),\boldsymbol\lambda^{(k)})
\end{equation}
where $f(\boldsymbol\Theta^{(k)}), \boldsymbol{\lambda}^{(k)}$ are the quantized weights and error variable at the $k^{th}$ iteration. The gradient of the loss function in Equation (\ref{equ:AL2}) w.r.t  $\boldsymbol\Theta$ is calculated as the following.

\begin{equation}\label{equ:gradient}
  \nabla L=\nabla\mathcal{F}_{ce}+\gamma(\boldsymbol{\Theta} - f(\boldsymbol\Theta^{(k)})+\boldsymbol{\lambda}^{(k)})
\end{equation}

It is found in practice that the quadratic term of the augmented Lagrangian of Equation (\ref{equ:AL2}) can dominate the loss function computation and lead to a local optimum. Following the extra-gradient method suggested in~\cite{Korpelevi-1976}, the solution to address this issue in this paper is to perform the gradient calculation by one additional step ahead to improve the convergence. 
\begin{equation}\label{equ:weightupdate}
  \begin{split}
  \bar{\boldsymbol{\Theta}} \leftarrow \boldsymbol\Theta^{(k)} - \eta_1\nabla L(\boldsymbol{\Theta})\\
  \boldsymbol{\Theta}^{(k+1)} \leftarrow \boldsymbol{\Theta}^{(k)} - \eta_2\nabla L(\bar{\boldsymbol{\Theta}})
\end{split}
\end{equation}

\noindent Here $\bar{\boldsymbol\Theta}$ represents the temporary variable used to store the intermediate updated parameters, and $\eta_1$ and $\eta_2$ are separate learning rates that are empirically set as $0.02$ and $0.001$ throughout the experiments of this paper.

\vspace{-1em}
\subsection{Quantization variables update}

The quantization variables including the scaling factor $\alpha$ in a globally shared quantization table $\{0, \pm\alpha,\dots, \pm \alpha (2^{n-1}-1)\}$, and the corresponding quantized parameters derived using Equation (17) can be solved by minimizing the following: 
 \begin{equation}
  \begin{split}
  &\min \limits_f ||\boldsymbol{\Theta}^{(k+1)}-f(\boldsymbol\Theta^{(k)})+\boldsymbol\lambda^{(k)}||_2^2 \\
  \Rightarrow&\min \limits_{\alpha, \textbf{V}} ||\boldsymbol{\Theta}^{(k+1)}+\boldsymbol\lambda^{(k)}-\alpha^{(k)}\boldsymbol{V}_\alpha^{(k)}||^2
\end{split}
\end{equation}
\noindent where alternating updates of the scaling factor estimate $\alpha^{(k)}$ and the associated quantized model parameters ${\bm V}_{\alpha}^{(k+1)}$ are performed via an inner-loop within the current $(k+1)^{th}$ outer iteration. 

The following algorithm shows the details of such inner-loop update. In all the experiments of this paper, the maximum number of inner-loops is set as 20 and practically found sufficient to ensure convergence.
\begin{algorithm}[h]
\SetAlgoLined
\KwResult{$\alpha^{(k+1)}$, $\boldsymbol{V}_{\alpha}^{(k+1)}$}
 $j=0$\;
 $\alpha_0^{(k+1)}=\alpha^{(k)}$\;
 $\boldsymbol{V}_{0, \alpha}^{(k+1)}= \arg \min\limits_{\boldsymbol{V}}|\boldsymbol{\Theta}^{(k+1)} - \alpha_0^{(k+1)} {\boldsymbol{V}} |$ \;
 \While{$j<20$ \&\& convergence is not reached}{
  $\alpha_{j+1}^{(k+1)} = \frac{(\boldsymbol{\Theta}^{(k+1)}+\boldsymbol\lambda^{(k)})^\top\boldsymbol{V}_{j,\alpha}^{(k+1)}}{\boldsymbol{V}_{j, \alpha}^{(k+1)\top}\boldsymbol{V}_{j,\alpha}^{(k+1)}}$\;
  $\boldsymbol{V}_{j+1,\alpha}^{(k+1)} = \arg \min\limits_{\boldsymbol{V}}|\boldsymbol{\Theta}^{(k+1)} - \alpha_{j+1}^{(k+1)} {\boldsymbol{V}}|$\;
  $j=j+1$\;
 }
 $\alpha^{(k+1)}=\alpha_j^{(k+1)}$\;
 $\boldsymbol{V}_{\alpha}^{(k+1)}=\boldsymbol{V}_{j,\alpha}^{(k+1)}$
 \caption{Quantization variables update inner loop}
\end{algorithm}

The minimization operation of Equation (24) aims to find the corresponding quantized model parameters given the full precision parameter updates in Section IV-A. As it is non-trivial to update both the scaling parameters $\alpha^{(l)}$ of the quantization table of Equation (18) and the quantized parameters derived using Equation (17) at the same time, an alternating estimation procedure is used here in the inner-loop of Algorithm 1, to produce interleaving updates of the scaling parameters $\alpha^{(l)}$ when fixing the current quantized parameters ${\bf V}_\alpha^{(k)}$, and vice versa when updating the quantized parameters ${\bf V}_\alpha^{(k)}$,  while keeping $\alpha^{(l)}$ unchanged. Intuitively the resulting update of $\alpha^{(l)}$ accounts for the change in the dynamic range of model parameters before and after quantization.

\vspace{-1em}
\subsection{Quantization error update}
The Lagrange multiplier variable $\boldsymbol\lambda$, now encoding the accumulated quantization errors computed at each iteration, is updated as
\begin{equation}\label{tml}
 \boldsymbol\lambda^{(k+1)} = \boldsymbol\lambda^{(k)} + \boldsymbol{\Theta}^{(k+1)}-f(\boldsymbol{\Theta}^{(k+1)})
\end{equation}

In all experiments of this paper the scaling factors $\alpha^{(l)}$ in Equation (18) are initialized to 1.0 before the ADMM update is performed. 

The above ADMM estimation of quantized neural network LMs can be executed iteratively until convergence measured in terms of validation data perplexity. In practice, a maximum number of 20 ADMM iterations was used throughout all the experiments of this paper to obtain convergence for all quantized LSTM-RNN and transformer LMs, as will be shown later in the convergence speed analysis of the following Section VI of experiments.

\vspace{-0.5em}
\section{Mixed Precision Quantization}
This section presents three approaches to automatically learn the optimal local precision settings previously introduced in Equation (18) for DNN quantization. The first two minimizes the performance sensitivity to low-bit quantization. They are measured using either the KL divergence between full precision and quantized LMs, or the log-likelihood curvature with respect to quantization error.  The third approach uses a mixed precision differentiable neural architecture search technique.  Examples of their application to Transformer and LSTM-RNN based LMs are also given.

\vspace{-1em}
\subsection{KL Divergence Based Mixed Precision Quantization}
The ultimate goal for any DNN quantization task, including the neural network LMs considered in this paper, is to obtain a “lossless” model compression such that the distance between the distribution embodied by the original full precision LM and that of the quantized model must be minimized. This requires the relative entropy, or equivalently the Kullback-Leibler (KL) divergence between full precision and quantized neural network LMs to be minimized. As a special case, the KL divergence is zero when a lossless compression is achieved, so that the same LM distribution is preserved after quantization. The use of KL divergence based distance metrics in early researches led to widely adopted back-off $n$-gram LM pruning techniques~\cite{Andreas-1998,Kristie-1996}.

In this paper, the KL divergence between the probability distributions obtained from a full precision NNLM and its quantized counterpart using a particular mixed precision setting is used to measure the resulting performance sensitivity. Taking a $L$-layer Transformer LM for example, for any quantization $f(\cdot)$ being applied to the full precision parameters ${\bf \Theta}$, the KL divergence based quantization sensitivity measure is computed over the training data of $N_w$ words as,
\vspace{-0.5em}
\begin{align}
  & \Omega^{\rm{KL}} = \sum_{i=1}^L\Omega^{\rm{KL}}_i = \sum_{i=1}^L D_{\rm{KL}}(P(\bm{\Theta_i})||P(f_{n_i}(\bm{\Theta_i})))\\
  &\mathmakebox[\displaywidth][r]{
  =\sum_{i=1}^{L} \sum_{t=1}^{N_w}P(w_t | w_{t-1}, h_{t-1}, {\bf \Theta}_i)\ln{\frac{P(w_t | w_{t-1}, h_{t-1}, {\bf \Theta }_i)}{P(w_t | w_{t-1}, h_{t-1}, f_{n_i}({\bf \Theta}_i)}}}\notag
 \end{align}
\noindent where ${\bf \Theta}_i$ denote the full precision parameters of the $i^{th}$ layer, and $f_{n_i}({\bf \Theta}_i)$ its associated $n_{i}$-bit quantized parameters given a particular local precision bit width $n_i$.

Given a target average quantization precision such as 2-bit, the local quantization bit widths used in each layer should be selected such that the total performance sensitivity in Equation (26) is minimized while satisfying the target model size constraint. However, directly evaluate the combined KL divergence metric for all possible mixed precision local quantization settings leads to a very large number of possible systems to be considered. For example, choosing among 4 different precision settings, 1-bit, 2-bit, 4-bit and 8-bit, across all 6 layers of a Transformer LM, produces a total of $4^6=4096$ mixed precision quantized Transformer LMs to be evaluated in terms of KL divergence against the full precision model. 

In order to address this scalability issue, a practical implementation adopted in this paper is based on a divide and conquer approach. The key information required to compute the sensitivity measure in Equation (26) is the local KL divergence metric, $\Omega_{i}^{KL}$, for example, associated with the $i^{th}$ Transformer LM layer. In our implementation, a set of prototype transformer LMs quantized using uniform precision, for example 1-bit, 2-bit, 4-bit and 8-bit are trained off-line first via ADMM optimization of Section IV. The performance sensitivity in Equation (26) can then be computed first locally for each layer by replacing the full precision parameters using each of the above four possible quantization choices at that layer only, before taking the sum to produce the combined net KL divergence metric when a particular mixed precision based local quantization configuration is applied across the entire Transformer LM. 

In order to further improve the efficiency when computing the KL divergence based sensitivity measure in Equation (26), a very small number of training data samples, as little as the data of one single mini-batch (batch size set as 32 throughout this paper) can be used, and adopted in all experiments of this paper. In practice this was found to produce quantized LM perplexity comparable to that obtained by computing the KL metric over the entire training data set. An ablation study on the relationship between the amount of Switchboard training data used to compute the KL divergence metric and determine the resulting mixed precision for Transformer LM quantization and the resulting LM's perplexity performance is shown in Table~\ref{tab:admm-ppl}. In order to further ensure efficiency, the resulting mixed precision quantized LMs using varying automatically learned local precision settings together with quantized parameters inherited from uniform precision models of different bit-widths are fine-tuned, rather than retrained from scratch. In practice, this was found to produce performance comparable to re-training them from scratch after determining the precision settings, as is illustrated in the example contrast of Table~\ref{tab:re-train} for 2-bit and 4-bit KL mixed precision quantized Transformer LMs on the Switchboard data.

An example application of KL divergence based mixed precision quantization of Transformer and LSTM-RNN LMs are shown in Figure 1 and 2 respectively.

\vspace{-1em}
\begin{table}[h]
\caption{\emph{Perplexity performance of Switchboard data trained average 2-bit quantized Transformer LMs with their local precision setting learned using the KL divergence metric of Equation (26), or the curvature based sensitivity metric of Equation (27). These two metrics were computed using either one single randomly drawn mini-batch (batch size 32), a randomly drawn 50\% subset, or all the training the data.}}
    \centering \begin{tabular}{c|c|c|c|ccc}
    \toprule
        \multirow{2}{*}{\textbf{LM}} & \textbf{quant.}  & \textbf{quant.} & \multirow{2}{*}{\textbf{\#bit}} &  \multicolumn{3}{c}{\textbf{PPL}} \\
        & \textbf{estim.} & \textbf{method} & & 1 & 50\% & 100\% \\ \midrule
        \multirow{3}{*}{Transformer} & - & - & 32 & - & -& 40.7 \\ \cline{2-7}
          & \multirow{2}{*}{ADMM} & KL &  1.9 & 46.0 & 46.3 & 45.9 \\
         &  & Hes & 1.9 & 46.4 & 46.0 & 46.2 \\ \bottomrule
    \end{tabular}
    \label{tab:admm-ppl}
    \vspace{-0.5em}
\end{table}

\begin{table*}[h]
\caption{\emph{Performace contrast between 2-bit and 4-bit KL mixed precision quantized Transformer LMs on the Switchboard data constructed using either post precision learning model fine-tuning (LM 2, 4), or retraining from scratch (LM 1, 3).}}
    \centering 
    \resizebox{180mm}{13mm}{
    \begin{tabular}{c|c|c|c|c|c|cc|ccc|cc|c}
    \toprule
        \multirow{2}{*}{\textbf{ID}} & \multirow{2}{*}{\textbf{LM}} & \textbf{quant.}  & \textbf{train} & \multirow{2}{*}{\textbf{\#bit}} & \multirow{2}{*}{\textbf{PPL}} & \multicolumn{2}{c}{\textbf{eval2000}}&  \multicolumn{3}{c}{\textbf{rt02}}&  \multicolumn{2}{c|}{\textbf{rt03}} & \textbf{WER} \\
        & & \textbf{method} & \textbf{method} & & & swbd & callhm & swbd1 &  swbd2 & swbd3 & fsh & swbd & avg. \\ \midrule
        1 & \multirow{4}{*}{Transformer}  & \multirow{4}{*}{KL} & re-train from scratch & \multirow{2}{*}{1.9} & 46.2 & 7.2 & 13.4 & 8.4 & 10.6 & 14.0 & 8.3 & 14.3 & 10.9    \\
        2 & &  & post-train fine-tuning & & 45.9  & 7.2 & 13.4 & 8.3 & 10.7 & 14.0 & 8.3 & 14.3 & 10.9 \\\cline{4-14}
        3 &   &  & re-train from scratch & \multirow{2}{*}{4} & 43.8 & 7.1 & 13.3 & 8.2 & 10.5 & 14.0 & 8.3 & 14.3 & 10.8   \\
        4 & &  & post-train fine-tuning &  & 44.0 & 7.1 & 13.3 & 8.2 & 10.5 & 14.0 & 8.3 & 14.2 & 10.8\\
        \bottomrule
    \end{tabular}}
    \label{tab:re-train}
\end{table*}

\begin{figure}[htbp]
  \setlength{\belowcaptionskip}{-1cm}
  \begin{center}
  \includegraphics[scale=0.17]{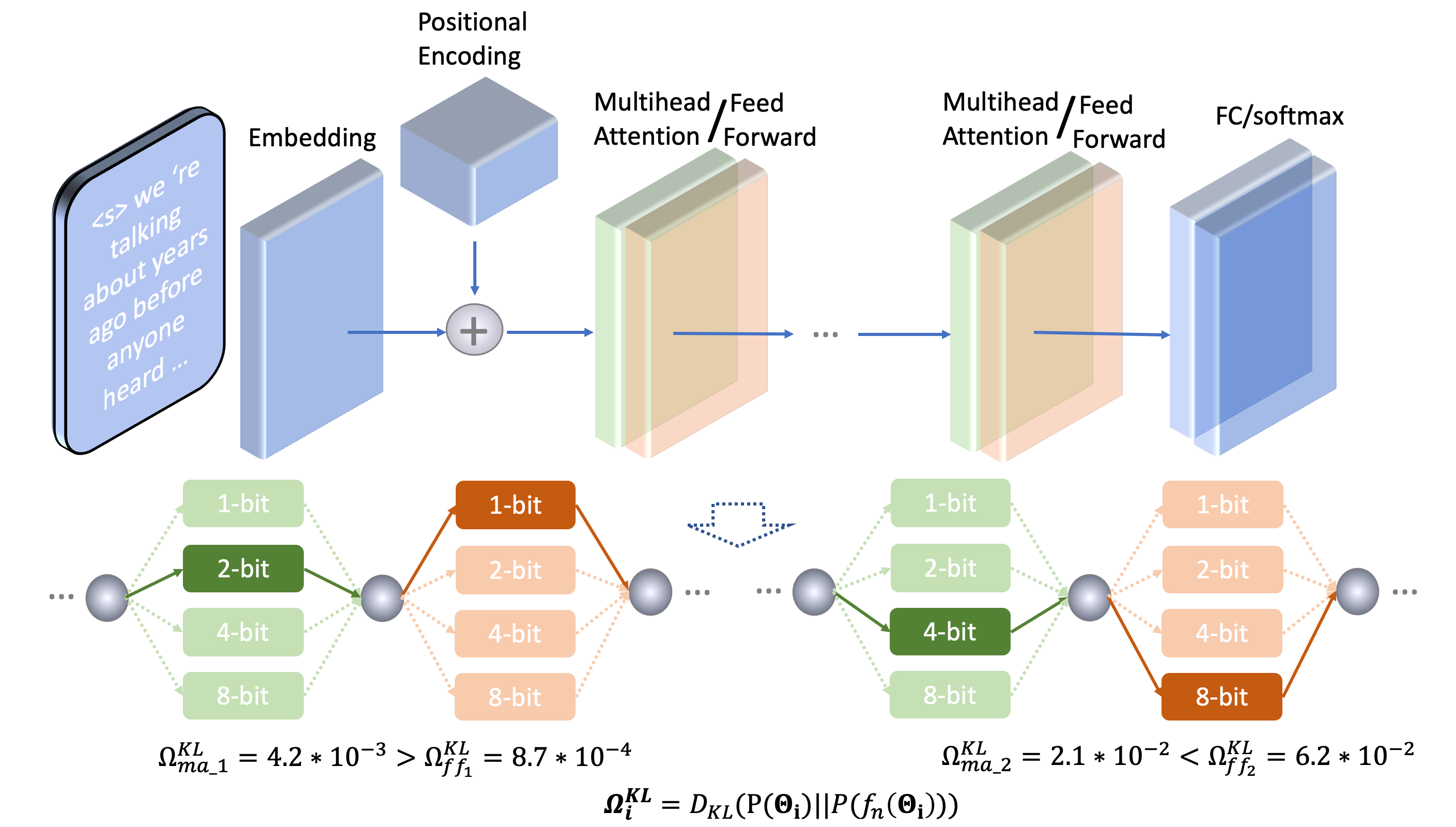}
  \caption{An example of  mixed precision quantization of a Transformer LM using KL-divergence based mixed precision quantization. For the first Transformer module positioned right after the embedding and position encoding layer, its multi-head attention layer (green) uses 2-bit quantization while its feed forward layer (orange) uses binary quantization precision, as determined by the KL-divergence based sensitivity measure.}
  \label{fig:per}
  \end{center}
  \vspace{-1.2em}
\end{figure}

\begin{figure}[htbp]
  \setlength{\belowcaptionskip}{-1cm}
  \begin{center}
  \includegraphics[scale=0.32]{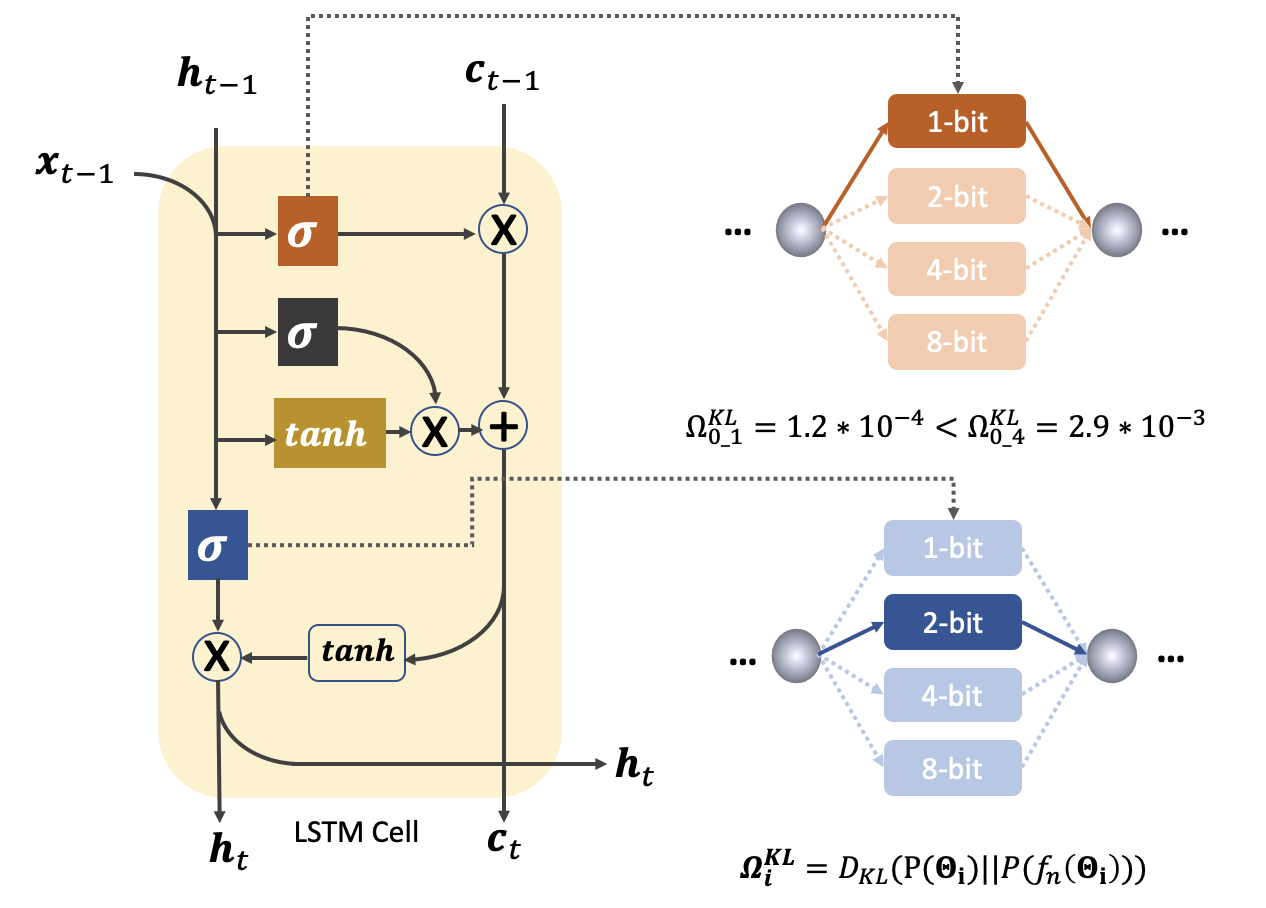}
  \caption{An example of auto-configured mixed precision quantization of a 2-layer LSTM-RNN LM with its forget, input and output Sigmoid gates $\sigma$ marked in orange, grey and dark blue, and cell gate $\tanh$ in brown respectively. The forget gate (orange) parameters use 1-bit quantization, while those of the output gate (dark blue) use 2-bit quantization, as determined by the KL-divergence based sensitivity measure.}
  \label{fig:per}
  \end{center}
  \vspace{-1.2em}
\end{figure}

\vspace{-1em}
\subsection{Curvature Based Mixed Precision Quantization}
The second approach to measure performance sensitivity to quantization examines the local training data log-likelihood curvature. Under mild assumptions such that the parameters of a DNN is twice differentiable and have converged to a local optimum, it has been shown in previous researches~\cite{ZDong-2019} that the performance sensitivity to quantization, when using a given precision setting, can be expressed as the squared quantization error further weighted by the parameter Hessian matrix trace. For any quantization $f(\cdot)$ being applied to the parameters $\boldsymbol{\Theta}$ of a $L$ layer  Transformer LM, the total performance sensitivity is given by the following sum of Hessian trace weighted squared quantization error. 
\begin{equation}
  \Omega^{\rm{Hes}} = \sum_{i=1}^L\Omega^{\rm{Hes}}_i = \sum_{i=1}^L Tr(\bm{H}_i)\cdot||f(\bm{\Theta}_i)-\bm{\Theta}_i||_2^2
\end{equation}

Intuitively for each cluster of weight parameters (for example of the same layer) to be quantized using a particular precision bit width, given the same amount of model parameter changes resulted from quantization, a smaller Hessian matrix trace indicates a lower performance sensitivity to quantization.

Given a target average quantization precision such as 2-bit, the local quantization setting, for example, used in each Transformer LM layer, should be selected such that the total performance sensitivity in Equation (27) is minimized while satisfying the target model size constraint. Similar to the KL divergence based mixed precision quantization, a set of prototype Transformer or LSTM-RNN LMs quantized using uniform precision, for example 1-bit, 2-bit, 4-bit and 8-bit, are trained off-line first using the ADMM optimization in Section IV. The log-likelihood curvature based performance sensitivity in Equation (27) can then be computed locally for each layer using each quantization choice before taking the sum to produce the combined net sensitivity measure for any mixed precision setting being considered. 
For larger LSTM-RNN or Transformer LMs containing up to hundreds of millions of parameters, and large DNNs in general, directly computing the Hessian matrix and its trace is computationally infeasible. In order to address this issue, an efficient stochastic linear algebra approach based on the Huchinson’s Algorithm~\cite{Avron-2011} is used to approximate the Hessian trace, 
\begin{equation}
  Tr(\bm{H})\approx \frac{1}{m}\sum_{i=1}^m \bm{z}_i^\top\bm{H}\bm{z}_i
\end{equation} 

\noindent where the expensive matrix multiplication between $\boldsymbol{H}$ and $\bm{z}_i$ can be avoided, and efficiently computed using Hessian-free approaches~\cite{ZDong-2019}. $\bm{z}_i$ is a random vector sampled from a Gaussian Distribution $\mathcal{N}(\bf{0}, \bf{1})$. Following the previous research reported in~\cite{ZDong-2019-2}, the maximum number of Hutchinson steps set as $m=50$ is found sufficient to obtain an accurate Hessian approximation for computing the curvature based quantization sensitivity of Equation (27), and used throughout the experiments of this paper. Again for efficiency, a small subset of the training data from a randomly drawn mini-batch (batch size 32) is used, in common with the previous KL divergence metric. Further analysis on the relationship between the sampled training data size and the resulting quantized Transformer LMs perplexity performance is shown in the last line of Table~\ref{tab:admm-ppl} for the Switchboard data.
An example application of the above curvature performance sensitivity based mixed precision quantization of a Transformer LM is shown in Figure 3. 
\begin{figure}[htbp]
  \setlength{\belowcaptionskip}{-1cm}
  \begin{center}
  \includegraphics[scale=0.17]{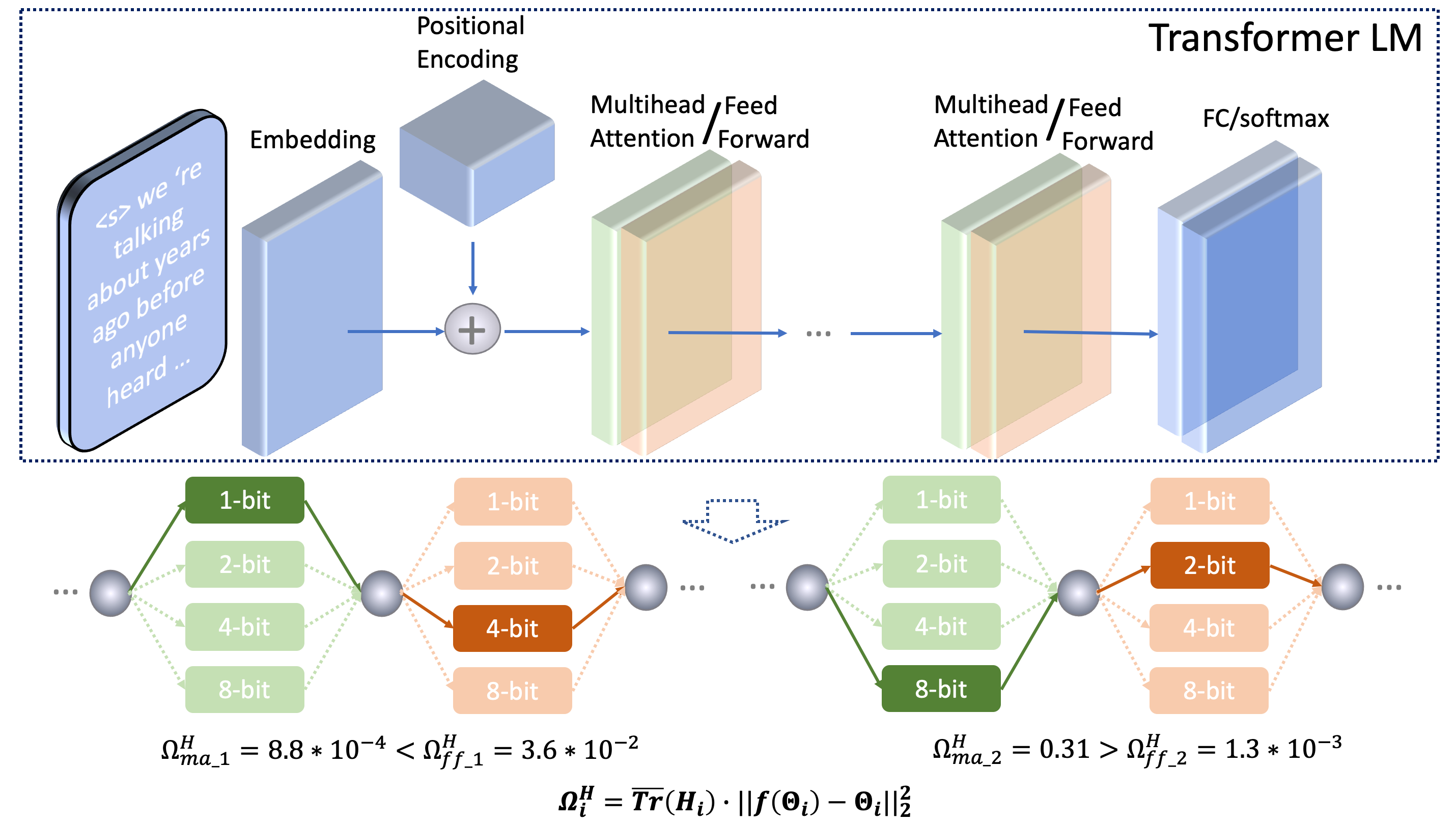}
  \caption{An example of auto-configured mixed precision quantization of a Transformer LM using curvature based sensitivity measure. For the first Transformer module positioned right after the embedding and position encoding layer, its multi-head attention layer (green) uses binary quantization while its feed forward layer (orange) uses 4-bit quantization precision, as determined by the Hessian-trace weighted quantization sensitivity measure. }
  \label{fig:per}
  \end{center}
  \vspace{-1.2em}
\end{figure}

\vspace{-1em}
\subsection{Architecture Search Based Mixed Precision Quantization}

\begin{figure}[htbp]
  \setlength{\belowcaptionskip}{-1cm}
  \begin{center}
  \includegraphics[scale=0.27]{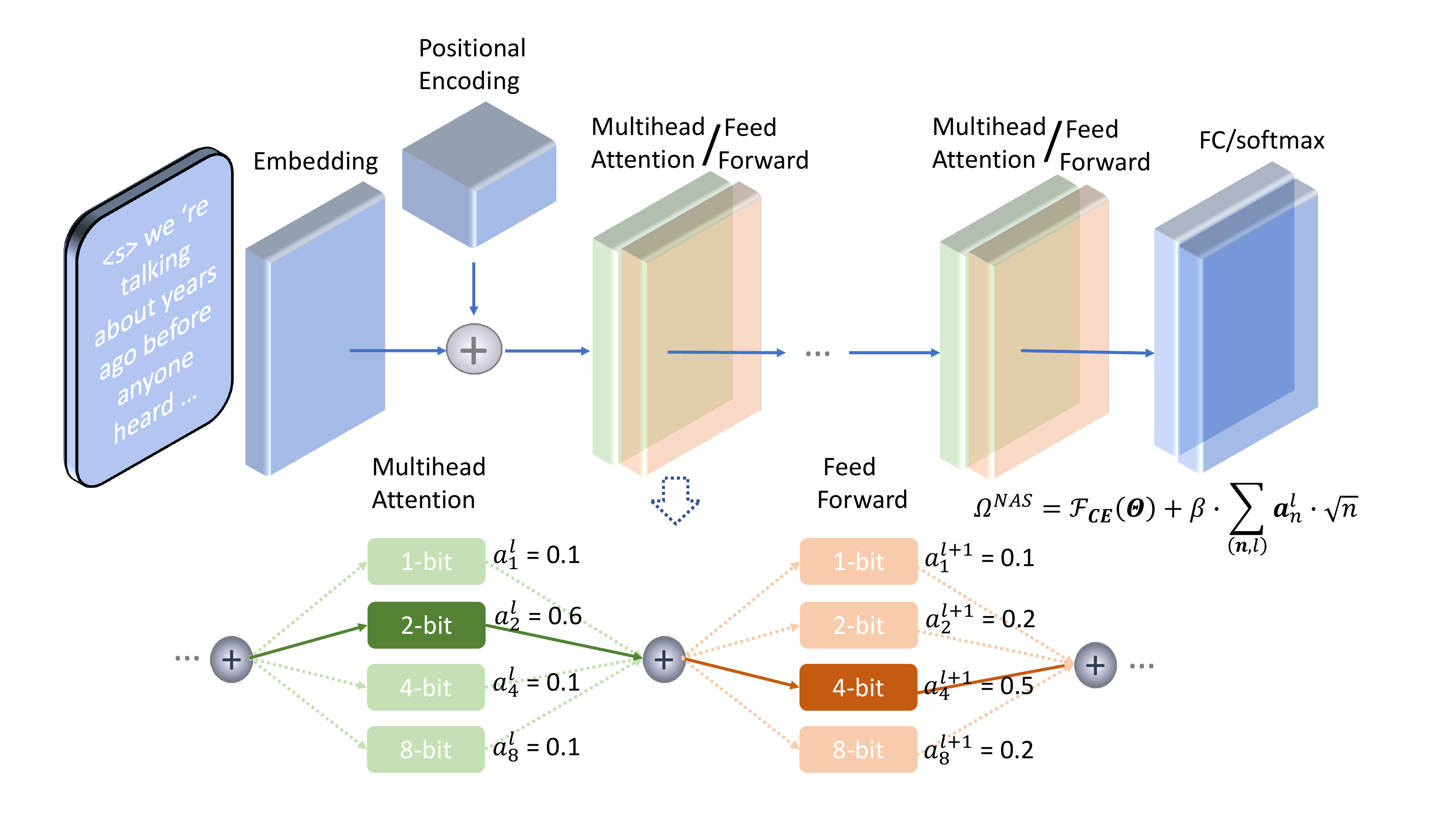}
  \caption{An example of auto-configured mixed precision quantization of a Transformer LM using mixed precision architecture search. For the first Transformer module, its multi-head attention layer is uses 2-bit quantization (green) given the associated selection weight of 0.6 while its feed forward layer uses 4-bit quantization precision (orange) given the associated selection weight of 0.5, as the 1-best choice selected from the mixed precision NAS super-network. }
  \label{fig:per}
  \end{center}
  \vspace{-1.2em}
\end{figure} 
The third solution to automatically learn the optimal local quantization precision settings is to use mixed precision based neural architecture search (NAS) approaches. Neural architecture search (NAS) techniques~\cite{Elsken-2019} can efficiently automate neural network structure designs that have been largely based on expert knowledge or empirical choice to date. Among existing NAS methods, differentiable neural architecture search (DARTS)~\cite{HLiu-2018,SXie2019,HCai-2019,HuS-2021,Hu-2020-CVPR} benefits from a distinct advantage of being able to simultaneously compare a very large number of candidate architectures during search time. This is contrast to earlier and more expensive forms of NAS techniques based on, for example, genetic algorithms~\cite{HCai-2020} and Reinforcement learning (RL)~\cite{Zoph-2017,Pham-2018}, where explicit system training and evaluation are required for a large number of candidate structures under consideration. 

Neural architecture search using DARTS is performed over an over-parameterized super-network containing paths connecting all candidate DNN structures to be considered. The search is transformed into the estimation of the weights assigned to each candidate neural architecture within the super-network. The optimal architecture is obtained by pruning lower weighted paths. This allows both architecture selection and candidate DNN parameters to be consistently optimized within the same super-network model. 

The key difference from conventional NAS tasks is that instead of selecting over heterogeneous neural building structures, for example, varying hidden layer left and right context offsets or projection layer dimensionality in LF-MMI trained time delay neural networks (TDNNs)~\cite{HuS-2021}, different quantized neural building blocks, for example, LSTM or Transformer LM modules of different bit-widths are considered. This crucial difference requires the associated mixed precision quantization super-network to be specially designed. Such super-network is constructed by first separately training transformer LMs using uniform precision, for example 1-bit, 2-bit, 4-bit and 8-bit, using ADMM optimization, before connecting these uniform precision quantized Transformer LMs at each layer, where the system specific activation outputs are linearly combined using a set of quantization precision selection weights as in the following equation:
\begin{equation}
    \boldsymbol{o}^l = \sum_{n\in N}\frac{exp(a_n^l)}{\sum_{}exp(a_n^l)}F(f_n(\boldsymbol{\Theta}^{l}), \boldsymbol{o}^{l-1})
\end{equation}
\noindent where $a^l_n$ is the architecture weights using $n$-bit quantization for the $l$-th cluster of weight parameters. $f_n(\cdot)$ represents $n$-bit quantization of the weight parameter $\boldsymbol{\Theta}^l$ and $F(\cdot)$ is the underlying layer's activation function. The set of possible quantization precision setting, $N=\{1,2,4,8\}$, is used throughout this paper. An example of such mixed precision Transformer super-network is shown in Figure 4. 

In order to avoid the trivial selection of the longest, most generous quantization bit width, these precision selection weights learning can be further constrained by a model complexity penalty term with respect to the number of bits retained after quantization, in order to obtain a target average quantization precision, for example, 2-bit,
\begin{equation}
    \Omega^{\rm{NAS}}=\mathcal{F}_{ce}(\boldsymbol{\Theta}) + \beta\sum_{(n, l)} a_n^l\cdot\sqrt{n} 
       \vspace{-0.5em}
\end{equation} 
\noindent where $\mathcal{F}_{ce}(\boldsymbol{\Theta})$ is the standard cross-entropy loss. 

\begin{table*}[h]
  \caption{\emph{Performance of the baseline full precision (LM 1), uniform precision quantized (LM 2-11) and mixed precision quantized RNNLMs with local layer or node level precision set either manually with equal bit-width (LM 12-15, 16-19) or automatically learned (LM 20-25) using KL, curvature (Hes) or NAS based mixed precision quantization methods of Section V on Switchboard NIST Hub5’00, RT02 and RT03. Equal weight interpolation of n-gram and neural LMs used in N-best restoring. All WER changes of no statistical significance (MAPSSWE, $\alpha=0.05$) over the full precision baseline (LM 1) are marked with "$\ast$". Offline, post-training quantized LMs’ performance are also shown (LM 6-9).}}
  \vspace{-0.7em}
  \label{tab:swbd-rnn}
  \centering
\resizebox{180mm}{42mm}{
  \begin{tabular}{c|c|c|c|c|c|cc|ccc|cc|c|c|c|c} 
      \toprule
          \multirow{2}{*}{\textbf{ID}}&  \textbf{quant.}&\textbf{param.} & \textbf{quant.}&\multirow{2}{*}{\textbf{\#bit}} & \multirow{2}{*}{\textbf{PPL}} & 
          \multicolumn{2}{c|}{\textbf{eval2000}}&
          \multicolumn{3}{c|}{\textbf{rt02}}&
          \multicolumn{2}{c|}{\textbf{rt03}}&\textbf{WER(\%)} &
      \textbf{model} & \textbf{comp.} & \textbf{eval. time}  \\  
          &\textbf{prec.}& \textbf{estim.} &\textbf{method}& & & swbd & callhm & swbd1 & swbd2 & swbd3 & fsh. & swbd & avg.& \textbf{size(MB)} & \textbf{ratio} & \textbf{(ms./word)}  \\ \midrule
0 & \multicolumn{3}{|c|}{4-gram} & - & 85.3 & 9.5 & 17.8 & 11.4 & 15.1 & 19.5 & 12.4 & 19.3 & 15.0 & - & - & - \\ \cline{1-2}\cline{3-17}
1 & \multicolumn{3}{|c|}{baseline (full precision)} & 32 & 40.7 & 7.1 & 13.2 & 8.2 & 10.7 & 13.8 & 8.4 & 14.1 & 10.8 & 183 & - & 0.226 \\ \cline{1-2}\cline{3-17}
2 & & & & 1 & 52.4 & 8.3 & 15.9 & 9.6 & 12.0 & 15.8 & 9.2 & 15.6 & 12.3 & 6.1 & 30.0 & 0.086 \\ \cline{1-1}
3 & & BP~\cite{Soudry-2014,Courbariaux-2015} & &  \boxed{2} & 50.1 & 8.0 & 15.1 & 9.3 & 11.6 & 15.4 & 9.1 & 15.4 & 12.0 & 11.8 & 15.5 & 0.140 \\ \cline{1-1}
4 & & (\emph{modified})&  & 4 & 49.7 & 7.8 & 14.5 & 8.7 & 11.3 & 15.1 & 8.9 & 15.0 & 11.6 & 23.5 & 7.9 &  0.142 \\ \cline{1-1}
5 &  &  & & 8 & 44.2 & 7.4 & 13.9 & 8.6 & 11.1 & 14.6 & 8.7 & 14.8 & 11.3 &  46.2 & 3.9 & 0.141 \\\cline{1-1}\cline{3-3}\cline{5-17}
6 &\emph{uniform}  &  & & 1 & 50.3 & 8.2 & 15.4 & 9.6 & 12.0 & 15.7 & 9.1 & 15.6 & 12.2 & 6.1 & 30.0 & 0.085  \\ \cline{1-1}
7 & prec. & Post-training~\cite{fasoli21_interspeech} & & 2 & 47.6 & 7.9 & 14.6 & 9.2 & 11.3 & 15.2 & 9.0 & 15.2 & 11.8 & 11.8 & 15.5 & 0.141 \\ \cline{1-1}
8 & & Offline Quant & & 4 & 47.1 & 7.8 & 14.3 & 8.7 & 11.2 & 14.9 & 8.8 & 15.0 & 11.5 & 23.5 & 7.9 & 0.142   \\ \cline{1-1}
9 & & & & 8 & 44.8 & 7.3 & 13.6 & 8.4 & 11.1 & 14.4 & 8.7 & 14.6 & 11.2 & 46.2 & 3.9 & 0.141  \\ \cline{5-17}\cline{1-1}\cline{3-3}
10 & &  ADMM & & 1 & 49.9 & 8.0 & 14.8 & 9.6 & 11.8 & 15.7 & 9.1 & 15.4 & 12.1 & 6.1 & 30.0 & 0.083 \\ \cline{1-1}
11 & & (\emph{global})& & 8 & 43.0 & 7.2 & 13.6 & 8.4 & 11.0 & 14.3 & 8.6 & 14.6 & 11.1 & 46.2 & 3.9 & 0.144\\ \cline{5-17}\cline{1-3}
12 & & ADMM & \emph{manual} & 1 &  49.4 & 7.8 & 14.2 & 9.4 & 11.7 & 15.5 & 8.8 & 15.1 & 11.8 &  7.6 & 24.1 & 0.102 \\ \cline{1-1}
13& & (\emph{multiple} & \emph{define} & \boxed{2}& 45.3 & 7.6 & 14.0 & 9.1 & 11.4 & 15.2 & 8.7 & 14.9 & 11.6 & 13.3 & 13.6 & 0.155 \\ \cline{1-1}
14 &  & \emph{eql. \#bit} & & 4 & 44.7 & 7.3* & 13.5* & 8.6 & 11.0 & 14.5 & 8.6* & 14.7 & 11.2 & 24.8 & 7.4 & 0.152  \\ \cline{1-1}
15 & & \emph{node}) & & 8 & 42.1 & 7.2* & 13.3* & 8.3* & 10.9* & 14.0* & 8.4* & 14.3* &  10.9* & 47.5 & 3.8 &  0.158 \\ \cline{1-1}\cline{5-17}\cline{3-3}
16 &\emph{mixed} & ADMM & &1 & 50.3 & 8.0 & 15.3 & 9.5 & 11.8 & 15.6 & 9.1 & 15.4 & 12.1 & 6.1 & 30.0 & 0.088 \\ \cline{1-1}
17 & prec. & (\emph{multiple} & &  \boxed{2}& 48.2 & 7.7 & 14.5 & 9.2 & 11.4 & 15.3 & 8.9 & 15.2 &  11.7 & 11.8 & 15.5 & 0.148 \\ \cline{1-1}
18 & & \emph{eql. \#bit} & &  4 & 46.3 & 7.6 & 14.0 & 8.6 & 11.2 & 14.7 & 8.8 & 14.9 & 11.4 & 23.5 & 7.9 & 0.148 \\ \cline{1-1}
19 & & \emph{layer)} & &  8 & 43.9 &  7.3* & 13.5* & 8.4* & 11.0* & 14.3 & 8.6* & 14.5 & 11.1* & 46.2 & 3.9 & 0.147 \\ \cline{1-1}\cline{3-17}
20 & &  & \multirow{2}{*}{Hes} & \boxed{1.9} & 46.2 & 7.5 & 14.1 & 8.8 & 11.1 & 14.5 & 8.9 & 15.0 &11.4 & 11.7 & 15.6 & 0.130 \\ \cline{1-1}
21 & & ADMM &  & 4 & 43.1 & 7.4 & 13.6 & 8.4 & 10.9 & 14.1 & 8.6 & 14.5 & 11.0 & 23.8 & 7.7 & 0.140 \\ \cline{1-1}\cline{4-17}
22 & & (\emph{layer level} & \multirow{2}{*}{KL} & \boxed{\textbf{1.9}}& \textbf{45.9} & \textbf{7.4*} & \textbf{13.9} & \textbf{8.5*} & \textbf{10.9*} & \textbf{14.2} & \textbf{8.7*} & \textbf{14.8} & \textbf{11.2} & \textbf{11.7} & \textbf{15.6} & \textbf{0.132}\\ \cline{1-1}
23 & & \emph{variable}) & & 4 & 41.9 & 7.3 & 13.4 & 8.3 & 10.8 & 14.0 & 8.5 & 14.4 & 10.9 & 23.7 & 7.7 & 0.139 \\ \cline{1-1}\cline{4-17}
24 & &  & \multirow{2}{*}{NAS} & \boxed{2.3} & 48.3 & 7.7 & 14.3 & 9.0 & 11.4 & 14.9 & 9.0 & 15.2 & 11.6 & 13.1 & 13.9 & 0.136 \\ \cline{1-1}
25 & &  &  & 4 & 46.8 & 7.6 & 14.1 & 8.5 & 11.2 & 14.6 & 8.7 & 14.7 & 11.3 & 23.4 & 7.8 & 0.141
\\ \bottomrule
      \end{tabular}
      }
          \vspace{-1em}
\end{table*}

\begin{table*}[h]
  \caption{\emph{Performance of the baseline full precision (LM 1), uniform precision quantized (LM 2-11) and mixed precision quantized Transformer LMs with local layer or node level precision set either manually with equal bit-width (LM 12-15, 16-19) or automatically learned (LM 20-25) using KL, curvature (Hes) or NAS based mixed precision quantization methods of Section V on SWBD NIST Hub5’00, RT02 and RT03. Equal weight interpolation of n-gram and neural LMs used in N-best restoring. All WER changes of no statistical significance (MAPSSWE, $\alpha=0.05$) over the full precision baseline (LM 1) are marked with "$\ast$". Offline, post-training quantized LMs’ performance are also shown (LM 6-9).}}
  \vspace{-0.7em}
  \label{tab:swbd-trans}
  \centering
\resizebox{180mm}{42mm}{
  \begin{tabular}{c|c|c|c|c|c|cc|ccc|cc|c|c|c|c} 
      \toprule
      \multirow{2}{*}{\textbf{ID}}&  \textbf{quant.}&\textbf{param.} & \textbf{quant.}&\multirow{2}{*}{\textbf{\#bit}} & \multirow{2}{*}{\textbf{PPL}} & 
      \multicolumn{2}{c|}{\textbf{eval2000}}&
      \multicolumn{3}{c|}{\textbf{rt02}}&
      \multicolumn{2}{c|}{\textbf{rt03}}&\textbf{WER(\%)} &
  \textbf{model} & \textbf{comp.} & \textbf{eval. time}  \\  
      &\textbf{prec.}& \textbf{estim.} &\textbf{method}& & & swbd & callhm & swbd1 & swbd2 & swbd3 & fsh. & swbd & avg.& \textbf{size(MB)} & \textbf{ratio} & \textbf{(ms./word)}  \\ \midrule
0 & \multicolumn{3}{|c|}{4-gram} & - & 85.3 & 9.5 & 17.8 & 11.4 & 15.1 & 19.5 & 12.4 & 19.3 & 15.0 & - & - & - \\ \cline{1-2}\cline{3-17}

1 & \multicolumn{3}{|c|}{baseline (full precision)} & 32 & 40.7 & 7.1 & 13.4 & 8.2 & 10.5 & 14.0  & 8.3 & 14.3 & 10.8 & 106 & - & 0.127  \\ \cline{1-2}\cline{3-17}
2 & &  & & 1 & 52.4 & 7.5 & 14.2 & 8.5 & 11.0 & 14.7 & 9.0 & 14.9 & 11.4 & 3.6 & 29.5 & 0.043  \\ \cline{1-1}
3 &  & BP~\cite{Soudry-2014,Courbariaux-2015} & & \boxed{2} & 50.1 & 7.5 & 14.1 & 8.5 & 11.0 & 14.6 & 8.9 & 14.8 & 11.3 &  7.2 & 14.7 & 0.059  \\ \cline{1-1}
4 &  & \emph{(modified)}  & & 4 & 49.7 & 7.4* & 13.9* & 8.4* & 10.8* & 14.4* & 8.7* & 14.6* & 11.0 & 13.8 & 7.7 & 0.063 \\ \cline{1-1}
5 &  & &  & 8 & 44.2 & 7.3* & 13.7* & 8.4* & 10.8* & 14.3* & 8.6* & 14.5* & 10.9 & 27.0 & 3.9 & 0.065 \\ \cline{1-1}\cline{3-3}\cline{5-17}
6 & \emph{uniform}&  & & 1 & 50.7 & 7.4 & 14.1 & 8.3 & 10.9 & 14.4 & 8.8 & 14.6 & 11.2 & 3.6 & 29.5 & 0.044 \\ \cline{1-1}\cline{1-1}
7 & prec. & Post-training~\cite{fasoli21_interspeech} & & 2 &  49.2 & 7.3 & 13.8 & 8.3 & 10.7 & 14.1 & 8.5 & 14.5 & 11.1 & 7.2 & 14.7 & 0.061 \\ \cline{1-1}\cline{1-1}
8 & & Offline Quant & & 4 & 47.3 & 7.2 & 13.7 & 8.3 & 10.6 & 14.0 & 8.4 & 14.4 & 10.9 & 13.8 & 7.7 & 0.059 \\ \cline{1-1}\cline{1-1}
9 & &  & & 8 & 45.0 & 7.2 & 13.5 & 8.2 & 10.6 & 14.0 & 8.3 & 14.4 & 10.9 & 27.0 & 3.9 & 0.062 \\ \cline{1-1}\cline{3-3}\cline{5-17}
10 & & ADMM & & 1 & 51.2 & 7.5 & 14.1 & 8.5 & 10.9 & 14.5 & 8.8 & 14.7 & 11.3 & 3.6 & 29.5 & 0.046 \\ \cline{1-1}\cline{1-1}
11 & & \emph{(global)} & & 8 & 44.0 & 7.2 & 13.6 & 8.4 & 10.8 & 14.2 & 8.5 & 14.5 & 10.9 & 27.0 & 3.9 & 0.062 \\ \cline{1-3}\cline{5-17}
12 & & ADMM & \emph{manual} &  1 & 49.4 & 7.3* & 13.8* & 8.3* & 10.8* & 14.2* & 8.6* & 14.5* & 11.0* & 4.3 & 24.7 & 0.059 \\ \cline{1-1}
13 & & (\emph{multiple} & \emph{define} & \boxed{2}& 45.3 & 7.2* & 13.5* & 8.3* & 10.6* & 14.0* & 8.4* & 14.4*  & 10.9* & 7.7 & 13.8 &  0.076 \\ \cline{1-1}
14 & & \emph{eql. \#bit} & & 4 & 44.7 & 7.1* & 13.5* & 8.2* & 10.5* & 14.0* & 8.3* & 14.3* & 10.8*  & 15.3 & 7.0 & 0.075 \\ \cline{1-1}
15 & & \emph{node}) & &  8 & 42.1 & 7.2* & 13.5* & 8.3* & 10.6* & 14.0* & 8.3* & 14.4* & 10.9* & 28.5 & 3.7 & 0.076 \\ \cline{1-1}\cline{3-3}\cline{5-17}
16 &\emph{mixed} & ADMM & & 1 & 50.3 & 7.3 & 14.0 & 8.3* & 10.8* & 14.3* & 8.7* & 14.6* & 11.2* & 3.7 & 29.5 & 0.048 \\ \cline{1-1}
17 & prec. &(\emph{multiple} &  & \boxed{2}& 48.2 & 7.2* & 13.6* & 8.3* & 10.6* & 14.0* & 8.5* & 14.4* & 11.0* & 7.2 & 14.7 & 0.070 \\ \cline{1-1}
18 & & \emph{eql. \#bit}&  & 4 & 46.3 & 7.2* & 13.6* & 8.2* & 10.7* & 14.0* & 8.3* & 14.3* & 10.9* & 13.8 & 7.7 & 0.072 \\ \cline{1-1}
19 & &\emph{layer)}&  & 8 & 43.9 & 7.2* & 13.5* &8.3* & 10.7* & 14.0* & 8.3* & 14.4* & 10.9* & 27.0 & 3.9 & 0.072 \\ \cline{1-1}\cline{3-17}
20 & &  & \multirow{2}{*}{Hes} & \boxed{1.9}& 46.2 & 7.2* & 13.5* & 8.3* & 10.7* & 14.0* & 8.3* & 14.4* & 10.9* & 7.0 & 15.1 & 0.062 \\ \cline{1-1}
21 & &  &  & 4 & 44.2 & 7.2 & 13.4 & 8.2 & 10.6 & 14.0 & 8.3 & 14.3 & 10.8 & 13.7 & 7.7 & 0.069 \\ \cline{1-1}\cline{4-17}
22 & & ADMM & \multirow{2}{*}{KL} & \boxed{\textbf{1.9}}& \textbf{45.9} & \textbf{7.2*} & \textbf{13.4*} & \textbf{8.3*} & \textbf{10.7*} & \textbf{14.0*} & \textbf{8.3*} & \textbf{14.3*} & \textbf{10.9*} & \textbf{7.0} & \textbf{15.1} & \textbf{0.064} \\ \cline{1-1}
23 & & (\emph{layer level} &  & 4 & 44.0 & 7.1 & 13.3 & 8.2 & 10.5 & 14.0 & 8.3 & 14.2 & 10.8 & 13.7 & 7.7 & 0.070 \\ \cline{1-1}\cline{4-17}
24 & & \emph{variable}) & \multirow{2}{*}{NAS} & \boxed{2.3} & 48.3 & 7.3* & 13.7* & 8.4* & 10.8* & 14.0* & 8.4* & 14.5* & 11.1* & 7.5 & 14.2 & 0.065 \\ \cline{1-1}
25 & &  &  & 4 & 45.8 & 7.2 & 13.5 & 8.2 & 10.6 & 14.1 & 8.3 & 14.3 & 10.9 & 13.4 & 7.9 & 0.068
\\ \bottomrule
      \end{tabular}
      }
          \vspace{-1.2em}
\end{table*}

\begin{table*}[h]
  \caption{\emph{Performance of the baseline full precision (LM 1), uniform precision quantized (LM 2-7) and mixed precision quantized RNNLMs with local layer or node level precision set either manually with equal bit-width (LM 8-11, 12-15) or automatically learned (LM 16-21) using KL, Hes or NAS based mixed precision quantization methods of Section V on AMI dev and eval sets of ihm, mdm8 and sdm1 conditions. Equal weight interpolation of n-gram and neural LMs used in N-best restoring. All WER changes of no statistical significance (MAPSSWE, $\alpha=0.05$) over the full precision baseline (LM 1) are marked with "$\ast$"}}
  \label{tab:ami-rnn}
  \centering
\resizebox{172mm}{40mm}{
  \begin{tabular}{c|c|c|c|c|c|cc|cc|cc|c|c|c} 
      \toprule
          \multirow{2}{*}{\textbf{ID}}& \textbf{param.} & \textbf{quant.}& \textbf{quant.}&\multirow{2}{*}{\textbf{\#bit}} & \multirow{2}{*}{\textbf{PPL}} & 
          \multicolumn{2}{c|}{\textbf{ihm}}&\multicolumn{2}{c|}{\textbf{mdm8}}&\multicolumn{2}{c|}{\textbf{sdm1}}&
       \textbf{model} & \textbf{comp.} & \textbf{eval. time} \\  
          &\textbf{prec.}& \textbf{estim.} & \textbf{method} & & & dev & eval & dev & eval & dev & eval & \textbf{size(MB)} & \textbf{ratio} & \textbf{(ms./word)}\\ \midrule
0 & \multicolumn{3}{|c|}{4-gram} & - & 93.6  & 18.1 & 18.4 & 33.5 & 35.3 & 34.2 & 40.5 & - & - & - \\ \hline
1 & \multicolumn{3}{|c|}{baseline (full precision)} & 32 & 57.6 & 16.4 & 16.2 & 27.7 & 30.8 & 29.9 & 34.7  &  248 & - &  0.290 \\ \hline
2 & & & & 1 & 78.6 & 17.2 & 16.9 & 32.5 & 34.1 & 33.9 & 40.1 &  8.2 & 30.2 & 0.108 \\  \cline{1-1}
3 &  & BP~\cite{Soudry-2014,Courbariaux-2015} & & 2 & 67.9 & 16.8 & 16.6 & 30.1 & 33.2 & 31.6 & 37.9 &  16.2 & 15.3 & 0.177 \\ \cline{1-1}
4 & \emph{uniform} & (\emph{modified)} & & 4 & 62.6 & 16.7 & 16.5 & 29.3 & 32.1 & 30.7 & 35.7 & 31.5 & 7.9 & 0.179 \\ \cline{1-1}
5 & prec. & & & 8 & 60.3 & 16.5* & 16.3* & 28.6 & 31.8 & 30.4 & 35.3 & 62.8 & 3.9 & 0.175 \\ \cline{1-1}\cline{3-3}\cline{5-15}
6 & & ADMM & & 1 & 74.2 & 17.0 & 16.8 & 31.2 & 33.7 & 32.9 & 39.6 & 8.2 & 30.2 & 0.110 \\ \cline{1-1}
7 & & (\emph{global}) & & 8 & 59.3 & 16.6* & 16.3* & 28.5 & 31.6 & 30.4 & 35.2 & 62.8 & 3.9 & 0.176 \\ \cline{1-3}\cline{5-15}
8 & & ADMM & \emph{manual} & 1 & 68.2 & 16.9 & 16.7 & 30.0 & 32.6 &  32.4 & 38.6 & 9.5 & 26.1 & 0.145 \\ \cline{1-1}
9 & & (\emph{multiple} & \emph{define} & 2 & 61.9 & 16.6* & 16.5* & 29.1 & 31.9 & 30.9 & 35.9 & 17.6 & 14.1 & 0.205 \\ \cline{1-1}
10 & & \emph{eql. \#bit} & & 4 & 58.5 & 16.4* & 16.3* & 28.6 & 31.2 & 30.5 & 35.2 & 33.0 & 7.5 & 0.207   \\ \cline{1-1}
11 & & \emph{node}) & & 8 & 57.9 & 16.4* & 16.3* & 28.4 & 31.1* & 30.1* & 34.9* & 64.5 & 3.8 & 0.204  \\ \cline{1-1}\cline{3-3}\cline{5-15}
12 &\emph{mixed} & ADMM & & 1 & 70.3 & 17.0 & 16.8 & 30.9 & 33.2 & 32.7 & 39.1 &  8.4 & 30.1 & 0.116 \\ \cline{1-1}
13 & prec. & (\emph{multiple} & & 2 & 63.1 & 16.7 & 16.6 & 29.4 & 32.1 & 31.2 & 36.2 &  16.2 & 15.3 & 0.189 \\ \cline{1-1}
14 & &  \emph{eql. \#bit} & & 4 & 59.2 & 16.6* & 16.4* & 28.9 & 31.7 & 30.9 & 35.4 & 31.5 & 7.9 & 0.190  \\ \cline{1-1}
15 & & \emph{layer}) & & 8 & 58.4 & 16.5* & 16.3* & 28.5 & 31.4 & 30.4 & 35.2 & 62.8 & 3.9 & 0.192 \\ \cline{1-1}\cline{3-15}
16 & &  & \multirow{2}{*}{Hes} & 1.9 & 60.7 & 16.5* & 16.4* & 28.6 & 31.8 & 30.8 & 35.3 & 16.0 & 15.5 & 0.165 \\ \cline{1-1}
17 & &  ADMM &  & 4.0 & 59.0 & 16.5* & 16.4* & 28.5 & 31.2 & 30.3* & 35.1 & 31.3 & 7.9 & 0.185 \\ \cline{1-1}\cline{4-15}
18 & & (\emph{layer}&  \multirow{2}{*}{KL} & 1.8 & 59.2 & 16.5* & 16.3* & 28.5 & 31.6 & 30.6 & 35.2 & 15.8 & 15.7 & 0.167 \\ \cline{1-1}
19 & & \emph{level} & & 3.8 & 58.8 & 16.5* & 16.3* & 28.4 & 31.3 & 30.2* & 35.0* & 30.8 & 8.1 & 0.183 \\ \cline{1-1}\cline{4-15}
20 & & \emph{variable}) & \multirow{2}{*}{NAS} & 2.0 & 63.6 & 16.8 & 16.5 & 28.9 & 32.0 & 31.1 & 35.9 & 16.2 & 15.3 & 0.171 \\ \cline{1-1}
21 & & & & 3.8 & 61.7 & 16.5* & 16.4* & 28.8 & 31.6 & 30.7 & 35.5 & 30.8 & 8.1 &  0.186
\\ \bottomrule
      \end{tabular}
       }
          \vspace{-1.5em}
\end{table*}

\begin{table*}[h]
  \caption{\emph{Performance of the baseline full precision (LM 1), uniform precision quantized (LM 2-7) and mixed precision quantized Transformer LMs with local layer or node level precision set either manually with equal bit-width (LM 8-11, 12-15) or automatically learned (LM 16-21) using KL, Hes or NAS based mixed precision quantization methods of Section V on AMI dev, eval sets of ihm, mdm8 and sdm1 conditions. Equal weight interpolation of n-gram and neural LMs used in N-best restoring. All WER changes of no statistical significance (MAPSSWE, $\alpha=0.05$) over the full precision baseline (LM 1) are marked with "$\ast$"}}
  \label{tab:ami-trans}
  \centering
\resizebox{172mm}{40mm}{
  \begin{tabular}{c|c|c|c|c|c|cc|cc|cc|c|c|c} 
      \toprule
      \multirow{2}{*}{\textbf{ID}}& \textbf{quant.} & \textbf{quant.}& \textbf{quant.}&\multirow{2}{*}{\textbf{\#bit}} & \multirow{2}{*}{\textbf{PPL}} & 
      \multicolumn{2}{c|}{\textbf{ihm}}&\multicolumn{2}{c|}{\textbf{mdm8}}&\multicolumn{2}{c|}{\textbf{sdm1}}&
   \textbf{model} & \textbf{comp.} & \textbf{eval. time} \\  
          &\textbf{prec.}& \textbf{estim.} & \textbf{method} & & & dev & eval & dev & eval & dev & eval & \textbf{size(MB)} & \textbf{ratio} & \textbf{(ms.word)}  \\ \midrule
0 & \multicolumn{3}{|c|}{4-gram} & - & 93.6  & 18.1 & 18.4 & 33.5 & 35.3 & 34.2 & 40.5  & - & - & - \\ \hline
1 & \multicolumn{3}{|c|}{baseline (full precision)}& 32 & 46.9 & 16.1 & 15.8 & 27.5 & 30.6 & 29.8 & 34.5 & 138 & - & 0.172  \\ \hline
2 & & & & 1 & 64.3 & 16.7 & 16.5 & 29.8 & 33.0 & 31.3 & 37.2 & 4.7 & 29.4 & 0.058 \\ \cline{1-1}
3 &  & BP~\cite{Soudry-2014,Courbariaux-2015} &  & 2 & 58.4 & 16.4 & 16.3 & 28.7 & 31.2 & 30.4 & 35.3 & 9.0 & 15.3 & 0.082 \\ \cline{1-1}
4 & \emph{uniform} & (\emph{modified}) & & 4 & 54.1 & 16.3* & 16.1* & 28.3 & 31.0 & 30.3 & 35.1 & 17.9 & 7.7 & 0.080 \\ \cline{1-1}
5 & prec. & & & 8 & 52.7 & 16.2* & 16.0* & 28.1 & 30.8 & 30.1 & 34.8 & 35.0 & 3.9 & 0.083 \\ \cline{1-1}\cline{3-3}\cline{5-15}
6 & & ADMM & & 1 & 59.9 & 16.6 & 16.5 & 29.7 & 32.6 & 31.1 & 37.0 & 4.7 & 29.5 & 0.062  \\ \cline{1-1}
7 & & (\emph{global}) & & 8 & 49.7 & 16.2* & 16.0* & 28.0 & 30.7* & 30.0* & 34.8* & 35.0 & 3.9 & 0.086 \\ \cline{1-3}\cline{5-15}
8 & & ADMM & \emph{manual} & 1 & 55.3 & 16.5 & 16.4 & 29.2 & 31.9 & 30.8 & 36.4 & 5.9 & 23.4 & 0.070 \\ \cline{1-1}
9 & & (\emph{multiple} & \emph{define} & 2 & 48.8 & 16.3* & 16.2* & 28.4 & 30.9 & 30.4 & 35.0 & 10.3 & 13.4 & 0.098 \\ \cline{1-1}
10 & & \emph{eql. \#bit} & & 4 & 49.0 & 16.1* & 15.9* & 27.9 & 30.7 & 30.0* & 34.7* & 19.4 & 7.1 & 0.098 \\ \cline{1-1}
11 & & \emph{node}) & & 8 & 48.3 & 16.1* & 15.8* & 27.8* & 30.7* & 29.9* & 34.6* & 36.3 & 3.8 & 0.096 \\ \cline{1-1}\cline{3-3}\cline{5-15}
12 &\emph{mixed} & ADMM & & 1 & 57.4 & 16.6 & 16.5 & 29.4 & 32.1 & 31.0 & 36.5 & 4.8 & 28.8 & 0.066  \\ \cline{1-1}
13 & prec. & (\emph{multiple} & & 2 & 52.9 & 16.3 & 16.3 & 28.5 & 31.1 & 30.4 & 35.1 & 9.1 & 15.2 & 0.092 \\ \cline{1-1}
14 & & \emph{eql. \#bit} & & 4 & 49.4 & 16.1* & 16.0* & 27.8* & 30.8* & 30.0* & 34.6* & 17.9 & 7.7 & 0.093 \\ \cline{1-1}
15 & & \emph{layer}) & & 8 & 48.3 &  16.1* & 15.9* & 27.9 & 30.7 & 30.0* & 34.7* & 35.0 & 3.9 & 0.092 \\ \cline{1-1}\cline{3-15}
16 & &  &  \multirow{2}{*}{Hes} & 1.8 & 48.1 & 16.1* & 16.0* & 28.0 & 30.8 & 30.1* & 34.7* & 8.7 & 15.9 & 0.075 \\ \cline{1-1}
17 & & ADMM &  & 4.0 & 49.3 & 16.1* & 15.9* & 27.8* & 30.7* & 29.9* & 34.6* & 17.8 & 7.8 & 0.086 \\ \cline{1-1}\cline{4-15}
18 & & (\emph{layer}& \multirow{2}{*}{KL} & 1.8 & 47.6 & 16.2*  & 16.1* & 28.1 & 30.9 & 30.1* & 34.8* & 8.7 & 15.9 & 0.074 \\ \cline{1-1}
19 & & \emph{level} & & 4.0 & 48.9 & 16.1* & 15.8* & 27.6* & 30.7* & 29.9* & 34.6* & 17.8 & 7.8 & 0.085 \\ \cline{1-1}\cline{4-15}
20 & & \emph{variable})& \multirow{2}{*}{NAS} & 1.9 & 49.8 & 16.3* & 16.2* & 28.3 & 31.0 &  30.3 & 35.0 & 9.0 & 15.3 & 0.077 \\ \cline{1-1}
21 & &  & & 3.9 & 49.7 & 16.2* & 16.1* & 28.1 & 30.7* & 30.0* & 34.9 & 17.7 & 7.8 & 0.087
\\ \bottomrule
      \end{tabular}
       }
          \vspace{-1.5em}
\end{table*}

\vspace{-0.5em}
\section{Experiments}
In this section the performance of mixed precision quantized LSTM-RNN and Transformer LMs are evaluated on two speech recognition systems both of which use state-of-the-art LF-MMI sequence trained hybrid time delay neural networks (TDNNs) acoustic models with factored weights and additional convolutional layers~\cite{Povey2016}. Speech perturbation based data augmentation and i-Vector based speaker adaptation are also employed. Modified KN smoothed 4-gram back-off LMs are used during the initial N-best list generation pass before various NNLMs are then applied in the following rescoring stage. In Section VI-A, the first set of experiments conducted on the Switchboard I (SWBD) corpus~\cite{Godfrey-SWBD} are presented. In Section VI-B another comparable set of experiments are carried out on the AMI meeting room data~\cite{Hain-AMI}.

All the LSTM-RMM LMs investigated in this paper consist of 2 LSTM layers while both the input word embedding and hidden layer sizes were set as 1024. All the Transformer LMs used in this paper contain 6 Transformer layers. The dimensionality of all query, key and value embedding and hidden vectors are set as 512 for each batch of data. All the mixed precision quantized LSTM-RNN LMs and Transformer LMs of this paper use layer or node level precision settings that are set either manually as equal bit-widths (1-bit, 2-bit, 4-bit or 8-bit), or automatically learned using the KL, curvature or NAS based mixed precision quantization methods of Sections V. All NNLMs were tained using a single NVIDIA Tesla V100 Volta GPU card. Statistical significance test was conducted at level $\alpha= 0.05$ based on matched pairs sentence segment word error (MAPSSWE) for recognition performance analysis. This is used to identify various “lossless” neural LM quantization configurations that incur no recognition performance degradation, or statistically significant word error rate (WER) increase.

The implementation used to evaluate the mixed precision quantization methods of this paper is exclusively based on the existing low-bit quantized precisions that are already natively supported by the NVidia Tesla V100 GPU. These include the use of the Boolean and masking operators to implement 1-bit quantization, and the INT8 data type used to implement 2, 4 and 8-bit quantization. In case of 2-bit and 4-bit quantization, extra padded bits of zero were also included.

\vspace{-1em}
\subsection{Experiments on Conversational Telephone Speech}
The Switchboard I telephone speech corpus we use consists of approximately 300 hours of audio data released by LDC (LDC97S62). Following the Kaldi toolkit~\cite{Povey2011} and its CNN-TDNN recipe\footnote{egs/swbd/s5c/local/chain/run\_cnn\_tdnn\_1a.sh}, LF-MMI trained CNN-TDNN acoustic models~\cite{Povey2016} with data augmentation and i-Vector adaptation~\cite{Dehak2011} were then built. The CNN-TDNN network consisted of 6 convolutional layers followed by 8 context-splicing TDNN layers with 1536 nodes per layer. A 160-dimensional factored linear projection was employed prior to affine transformation in each context-splicing layer other than the first one. ReLU activation functions were used, followed by batch normalization and dropout operations. Left and right context offsets of $\{-3,3\}$ were also used in the context-splicing layers. More detailed description fo the baseline system's acoustic modelling configuration can be found in~\cite{XXie2020}. The Switchboard NIST Hub5'00, RT02 and RT03 evaluation sets were used. A 30K word recognition lexicon was used. Various LSTM and Transformer LMs trained on the Switchboard and Fisher transcripts (LDC2004T19, LDC2005T19) were used to rescore the 4-gram LM produced N-best lists ($N=20$).

\noindent\emph{1) \textbf{Experiments on LSTM-RNN LMs}} using baseline full precision, various uniform and mixed quantization settings are shown in Table~\ref{tab:swbd-rnn}. All WER changes as the result of quantization that are of no statistical significance (MAPSSWE, $\alpha=0.05$) are marked by “$\ast$”. Several trends can be found here:
\begin{itemize}
  \item Among the uniform precision quantized LMs, the LSTM-RNN LMs of 1-bit and 8-bit precisions trained using the ADMM optimization of Section IV consistently outperform those of comparable bit-widths but built using the modified BP algorithm (LM 10 and 11 vs. LM 2 and 5 in Table~\ref{tab:swbd-rnn}). In particular, for 1-bit quantization, the ADMM optimization method produced WER reductions up to 1.1\% absolute on the {\bf callhm} data of the Hub5’00 set over the traditional modified BP algorithm (LM 10 vs. LM 2 in Table~\ref{tab:swbd-rnn}). The advantage of the ADMM algorithm in terms of convergence speed and training efficiency is further illustrated in left part of Figure 5.

  \item The mixed precision quantized LSTM LMs (LM 12-25 in Table~\ref{tab:swbd-rnn}) consistently outperform the uniform precision quantized models (LM 2-11). For example, given the same quantization precision at approximately 2-bit (compression ratio of 16 times over 32-bit full precision), a wide range of mixed precision quantized LMs (LM 13, 17, 20, 22, 24 in Table~\ref{tab:swbd-rnn}) produced statistically significant 0.3-0.8\% average WER reduction across three test data sets over with the comparable 2-bit uniform precision quantization baseline (LM 3).
  
  \item Note that the first two of these mixed precision quantized LSTM LMs (LM 13, 17 in Table~\ref{tab:swbd-rnn}) used multiple layer or node level locally applied quantization tables of the same bit-width manually set as 2-bit. As expected, a finer modelling granularity provided by the node level local quantization (LM 17) marginally outperformed the use of layer level (LM 13) by 0.1\% on average in WER, albeit with a smaller compression ratio of 13.6 vs. 15.5. Similar trends can be found when increasing the quantization precision from 2-bit to 8-bit\footnote{Considering the trade-off between WER and compression ratio for node and layer level local quantization precision settings, the subsequent experiments will focus on using layer level mixed precision quantization.}. 
  
  \item Among the three mixed precision quantization LSTM LMs with approximately 2-bit precision (LM 20, 22, 24 in Table~\ref{tab:swbd-rnn}), the KL quantized LM (LM 22) with a 1.9-bit average precision gives the lowest average WER of 11.2\%. These three mixed precision LMs' respective local selection of quantization bit-widths at different gates of various LSTM layers are shown in Figure 6(a)-6(c). Note that KL and curvature based quantization (LM 20, 22 in Table~\ref{tab:swbd-rnn}) both selected 8-bit precision for the 2nd LSTM layer's output gate parameters, indicating a larger performance sensitivity to quantization measured at the LSTM cells' output gate.
  
  \item Among all the quantized LSTM LMs marked with “$\ast$” in Table~\ref{tab:swbd-rnn} that incur no statistically significant WER increase against the full precision baseline (LM 1), the 1.9-bit KL mixed precision quantized model (LM 22) also produced the largest "lossless" compression ratio of 15.6.
\end{itemize}
\begin{figure*}
     \centering
     \subfigure[LSTM-RNN LMs]{\includegraphics[width=3.5in]{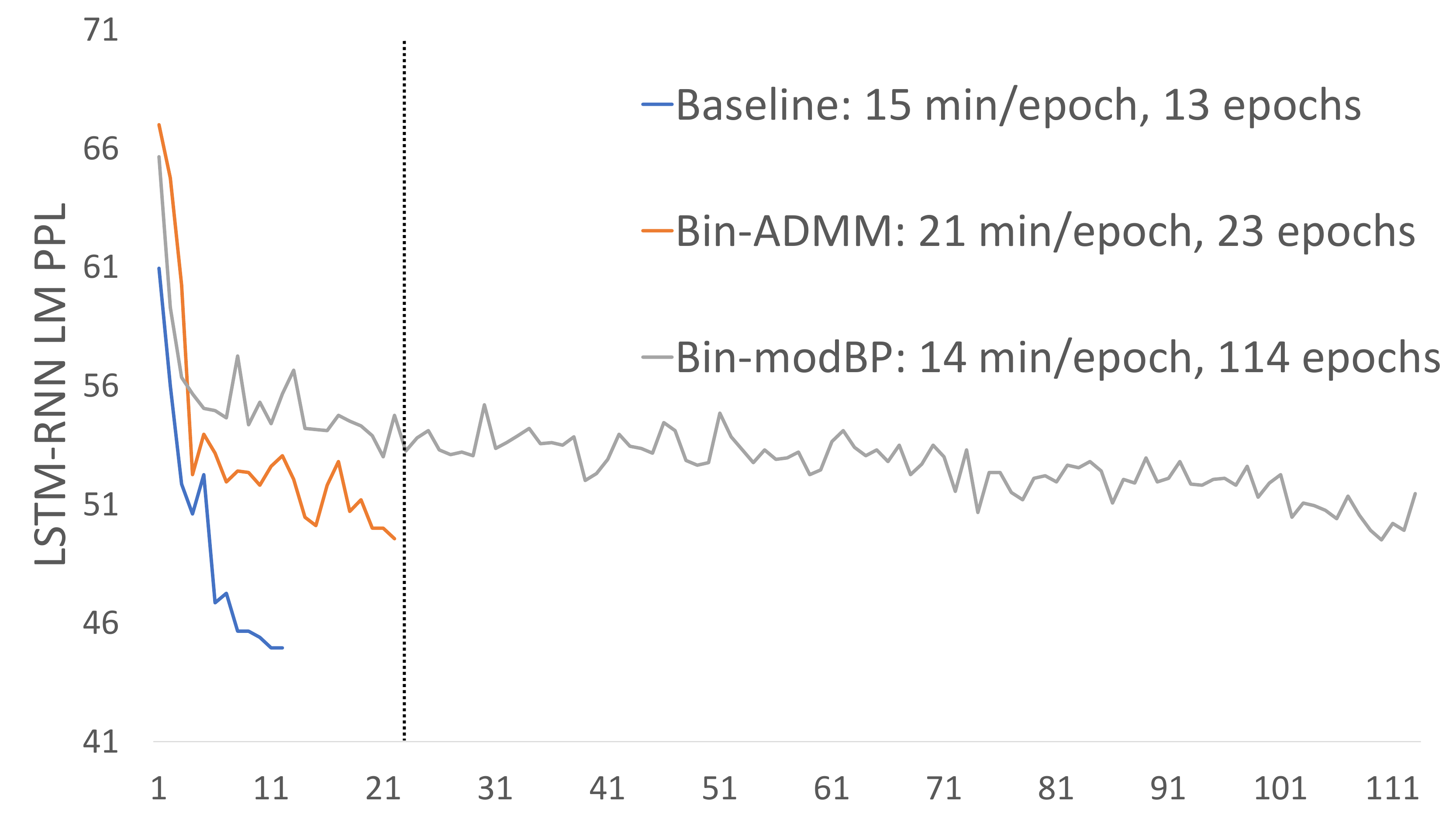}}
     \subfigure[Transformer LMs]{\includegraphics[width=3.5in]{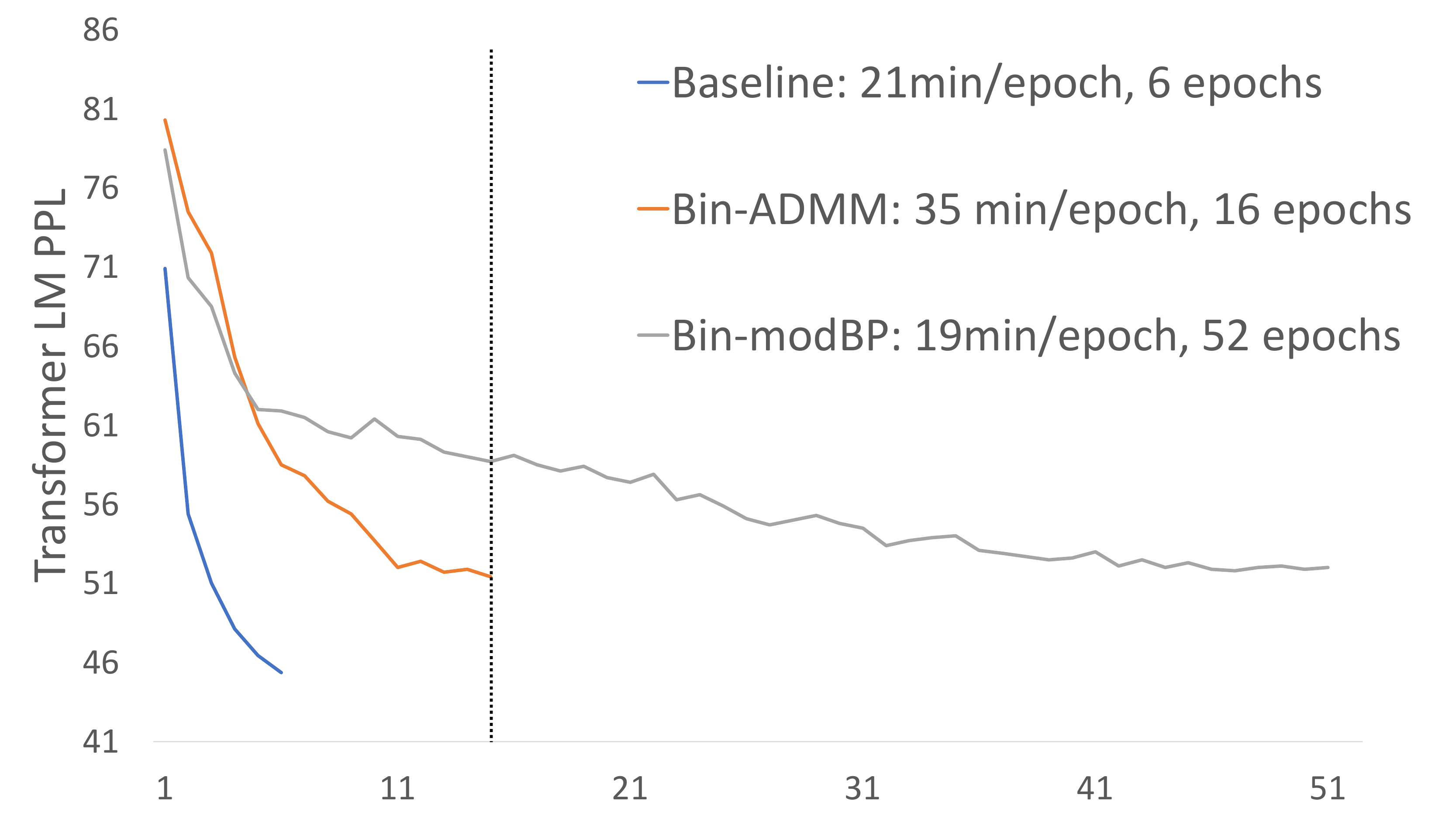}}
    \caption{Perplexity and convergence speed comparison on LSTM-RNN LMs (figure (a), showing LM 1, 2, 6 in Table~\ref{tab:swbd-rnn}) and Transformer LMs (figure (b), showing LM 1,2,6 in Table~\ref{tab:swbd-trans}) between baseline full precision models (\emph{(Baseline)}, binarized models trained using modified back-propagation (\emph{Bin-modBP})~\cite{Soudry-2014,Courbariaux-2015} and ADMM trained binarized models (\emph{Bin-ADMM}) on Switchboard validation data.}
    \label{fig:three graphs}
\end{figure*}
\noindent\emph{2) \textbf{Experiments on Transformer LMs}} designed using a comparable set of contrasts previously used in Table~\ref{tab:swbd-rnn} for LSTM-RNN LMs are shown in Table~\ref{tab:swbd-trans}, where a large number of quantized Transformer LMs featuring different forms of mixed quantization schemes (LM 12 to 25 in Table~\ref{tab:swbd-trans}, marked with “$\ast$”) produced lossless compression with no statistically significant WER changes relative to the baseline full precision model (LM 1). Several other trends that are similar to those found in Table~\ref{tab:swbd-rnn} on LSTM-RNN LMs are summarized below:

\begin{itemize}
  \item Among the uniform precision quantized LMs, the Transformer LMs of 1-bit and 8-bit precision trained using ADMM optimization of Section IV marginally outperform, or are comparable in performance to those two of same precisions using the modified BP algorithm (LM 10 and 11 vs. LM 2 and 5 in Table~\ref{tab:swbd-trans}). The advantage of the ADMM based Transformer LM estimation in terms of convergence speed and training efficiency is shown in right part of Figure 5.  
  
  \item The mixed precision quantized Transformer LMs (LM 8-18 in Table~\ref{tab:swbd-trans}) again consistently outperform the uniform precision quantized baselines (LM 2-11) across different lower bit-widths ranging from 1-bit, 2-bit to 4-bit. For example, given the same quantization precision at approximately 2-bit (compression ratio of 16 times), a wide range of mixed precision quantized Transformer LMs (LM 13, 20, 22 in Table~\ref{tab:swbd-trans}) produced statistically significant 0.4\% absolute WER reduction averaged across all data sets over with the comparable 2-bit uniform precision quantization LM (LM 3 in Table~\ref{tab:swbd-trans}). 
  
  \item Among the three mixed precision quantized Transformer LMs with approximately 1.9-bit precision (LM 20, 22, 24 in Table~\ref{tab:swbd-trans}), the KL and curvature based mixed precision quantized Transformer LMs (LM 22) with a 1.9-bit average precision produced the lowest average WER of 10.9\%, as well as the largest “lossless” compression ratio of 15.1 among all the quantized Transformer LMs marked with “$\ast$” in Table~\ref{tab:swbd-trans} that incur no statistically significant WER increase over the full precision baseline (LM 1 in Table~\ref{tab:swbd-trans}). Their respective local selection of quantization bit-widths at different sublayers within various Transformer layers are shown in Figure 6(e)-6(f), where an expected general trend is found that both the lower Transformer layers heavily tasked with de-noising the data, and the top Transformer layer used to immediately predict detailed word probabilities over a large vocabulary, require longer quantization precision settings (2-bit to 8-bit) than those middle Transformer layers (1-bit to 4-bit).    
  
  \begin{figure*}
\centering

\subfigure[KL quantized LSTM LM (avg. 2-bit)]{\includegraphics[width=0.31\linewidth]{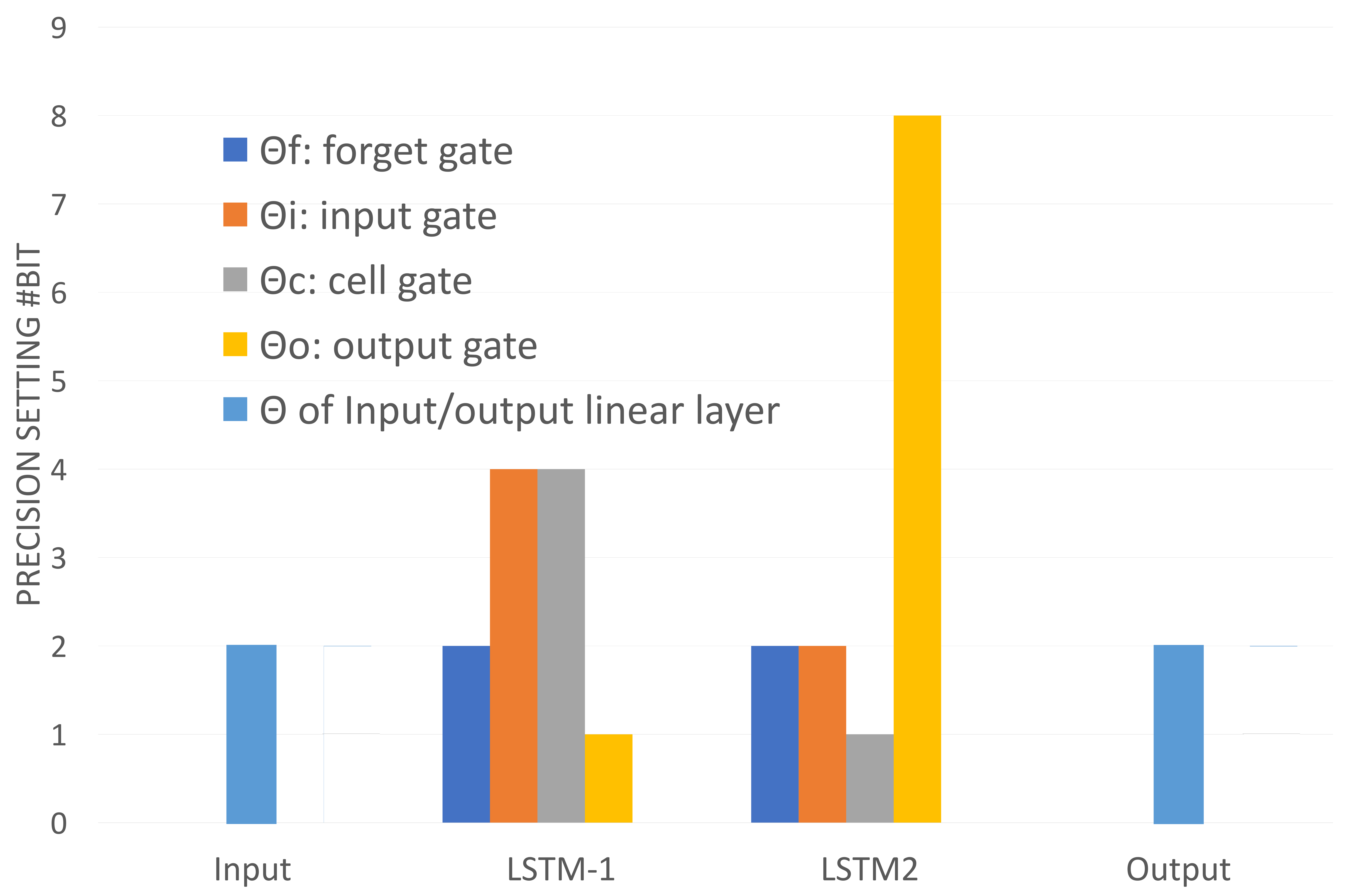}}
\subfigure[Curvature quantized LSTM LM (avg. 2-bit)]{\includegraphics[width=0.31\linewidth]{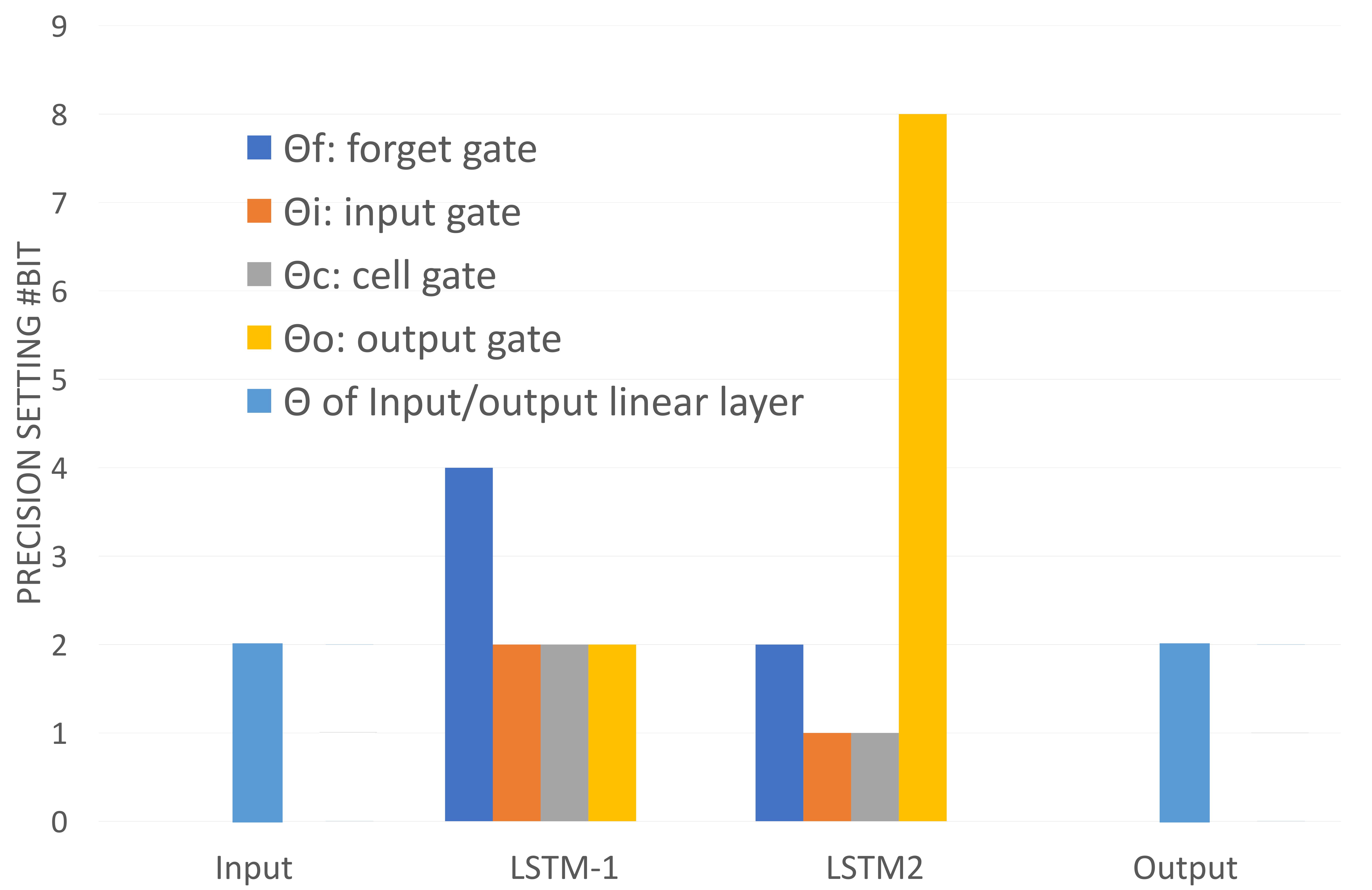}}
\subfigure[NAS quantized LSTM LM (avg. 2-bit)]{\includegraphics[width=0.31\linewidth]{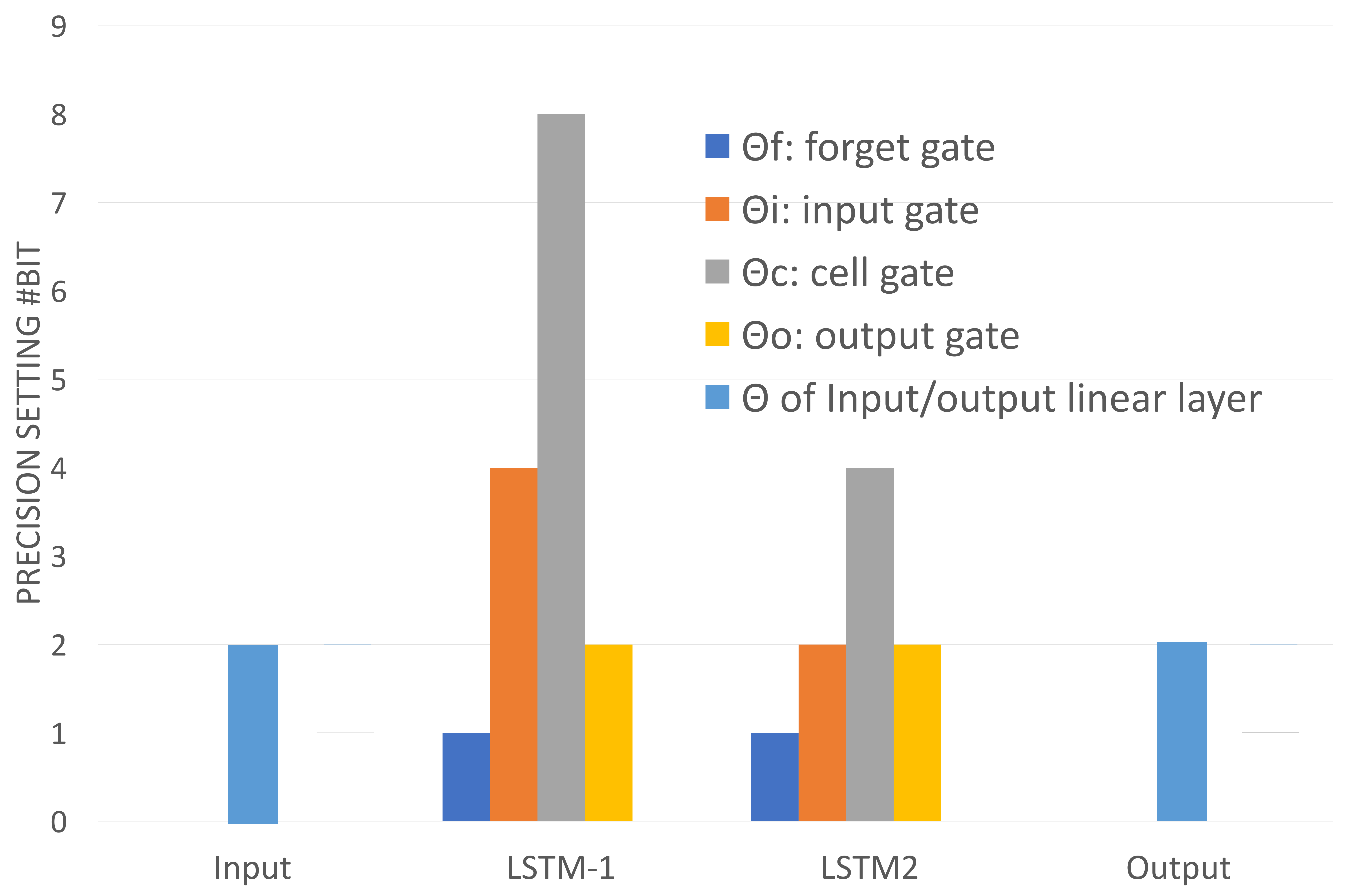}} \\
\subfigure[
KL quantized Transformer LM (avg. 2-bit)]{\includegraphics[width=0.31\linewidth]{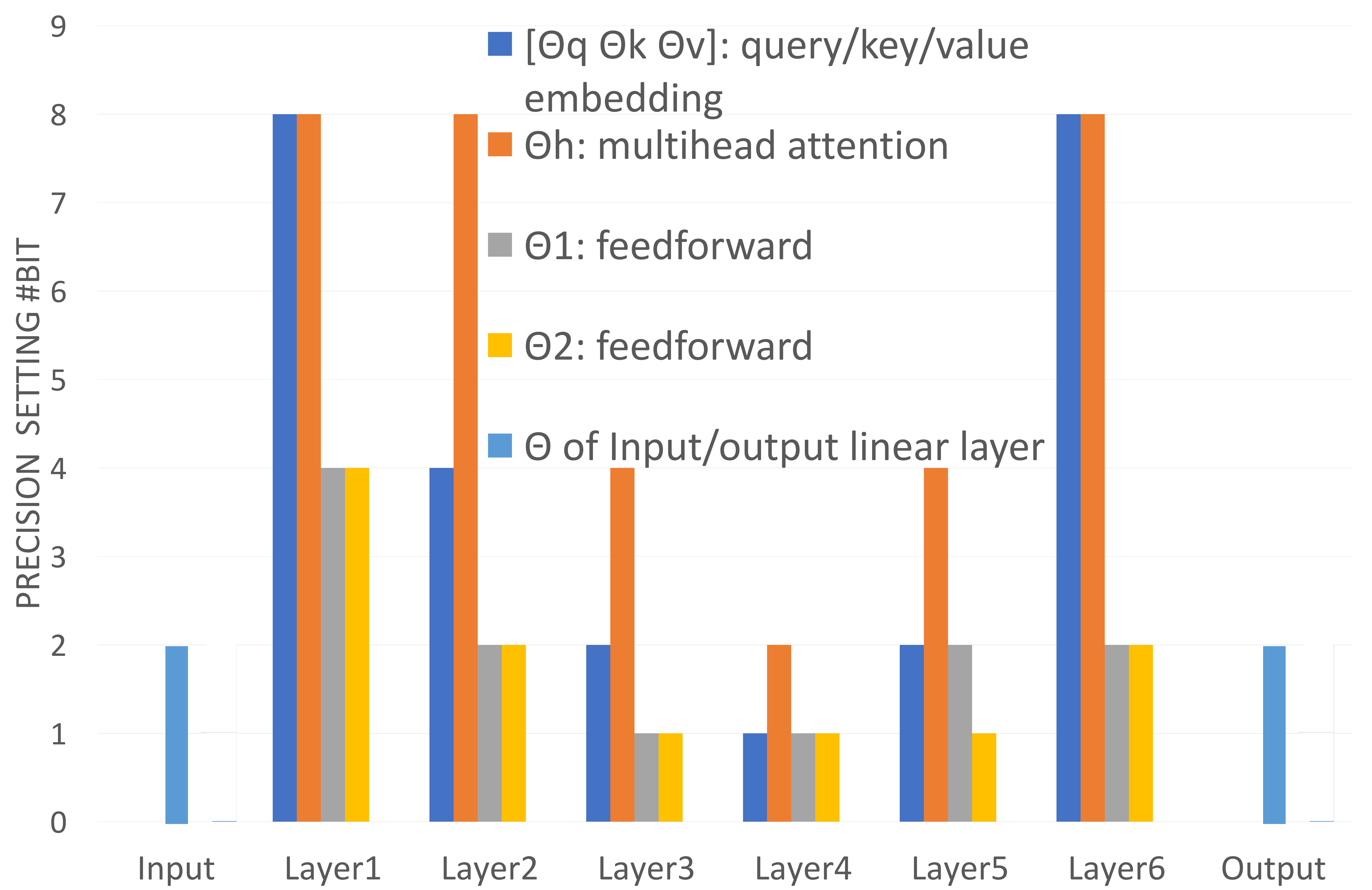}}
\subfigure[
Curvature quantized  Transformer LM (avg. 2-bit)]{\includegraphics[width=0.31\linewidth]{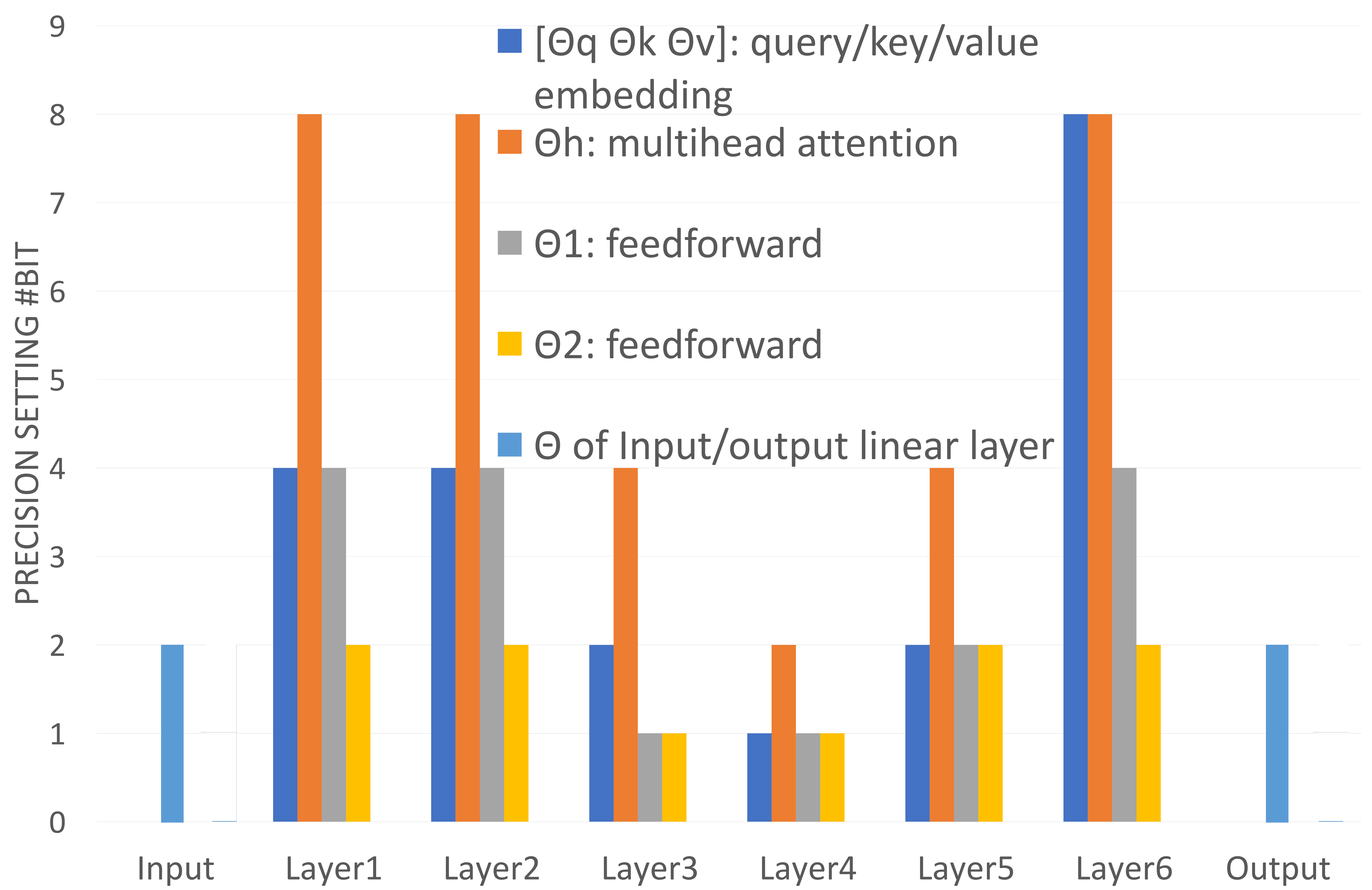}}
\subfigure[
NAS quantized Transformer LM (avg. 2-bit)]{\includegraphics[width=0.31\linewidth]{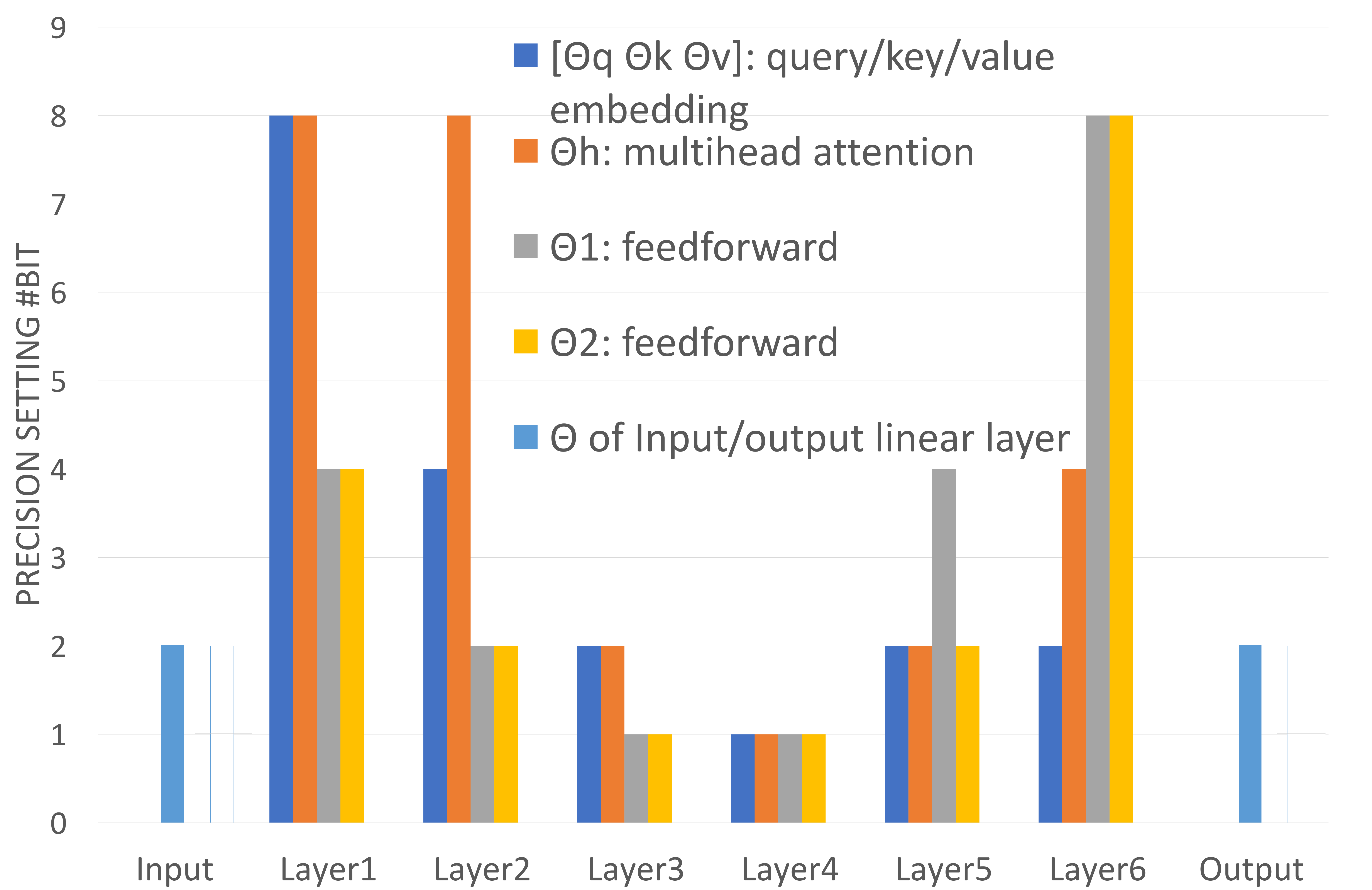}}
\caption{Number of bits used in local quantization of SWBD data trained neural LMs automatically derived using KL, curvature and NAS based mixed precision quantization of Section V for:
1) LSTM-RNN LMs ((a)-(c), also shown as LM 16-18 in Table~\ref{tab:swbd-rnn}) at their input/output linear layers (light blue), and individual forget (dark blue), input (orange), cell (grey) and output (yellow) gates inside LSTM layers; 2) Transformer LMs ((d)-(f), also shown as LM 16-18 in Table~\ref{tab:swbd-trans}) at their input/output linear layers (light blue), and 1st/2nd multihead attention sublayers (dark blue/orange), following 1st/2nd feedforward sublayers (grey/yellow) within each of 6 Transformer layers.}
\vspace{-1em}
\end{figure*}

  \item It is also worth noting the 4-bit ADMM node level locally quantized Transformer LM (LM 14 in Table~\ref{tab:swbd-trans}), and the 4-bit KL or curvature mixed precision quantized Transformer LM  (LM 21, 23 in Table~\ref{tab:swbd-trans}), both produced no WER degradation against the full precision model (LM 1), albeit with a smaller compression ratio of 7.0.
\end{itemize}

\vspace{-1em}
\subsection{Experiments on AMI Meeting Room Data}
The Augmented Multi-party Interaction (AMI) speech corpus consists of approximately 100 hours of audio data collected using both headset microphone and distant microphone arrays from the meeting environment. Following the Kaldi recipe\footnote{egs/ami/s5c/local/chain/run\_tdnn.sh}, three LF-MMI trained~\cite{Povey2016} acoustic models with speech perturbation based data augmentation and i-Vector based speaker adaptation~\cite{Dehak2011} were then constructed. The AMI 8.9-hour \textbf{dev} and 8.7-hour \textbf{eval} sets recorded under close talking microphone (\textbf{ihm}), single distant microphone (\textbf{sdm}) and multiple distant microphones (\textbf{mdm}) were used. A 47K word recognition lexicon was used. Various LSTM and Transformer LMs based on the same configurations as those on the Switchboard data in Table~\ref{tab:swbd-rnn} and~\ref{tab:swbd-trans} were trained using a mixture of text sources of 15M words including Fisher transcripts and 3 times of AMI transcriptions before being used to rescore the 3-gram LM produced N-best lists ($N=20$). All other experimental configurations remain the same as the Switchboard experiments of Section VI-A.

The mixed precision quantization experiments for RNNLMs and Transformer LMs on AMI meeting room data are shown in Table~\ref{tab:ami-rnn} and~\ref{tab:ami-trans} . The following trends similar to those previously found on the Switchboard data are again observed in Table~\ref{tab:ami-rnn} and~\ref{tab:ami-trans} for LSTM-RNN and Transformer LMs.

\begin{itemize}
    \item The mixed precision quantized LSTM and Transformer LMs (LM 8-21 in Table~\ref{tab:ami-rnn} and Table~\ref{tab:ami-trans}) consistently outperform the uniform precision quantized models (LM 2-7 in Table~\ref{tab:ami-rnn} and Table~\ref{tab:ami-trans}). For example, given the same quantization precision at approximately 4-bit (compression ratio of 8 times over 32-bit full precision), a wide range of mixed precision quantized LMs (LM 10, 14, 17, 19, 21 in Table~\ref{tab:ami-rnn} and Table~\ref{tab:ami-trans}) produced statistically significant WER reductions up to 0.9\% absolute (LM 19 vs. LM 4 on {\bf mdm} {\bf dev} in Table~\ref{tab:ami-rnn}) for LSTM LMs, and up to 0.7\% absolute WER reduction (LM 19 vs. LM 4 on {\bf mdm} {\bf dev} in Table~\ref{tab:ami-trans}) for Transformer LMs over the comparable 4-bit uniform precision quantization (LM 4).
    
    \item Among the six mixed precision quantization LSTM and Transformer LMs with approximately 4-bit precision (LM 17, 19, 21 in Table~\ref{tab:ami-rnn} and Table~\ref{tab:ami-trans}), both the KL quantized LSTM and Transformer LMs (LM 19 in Table~\ref{tab:ami-rnn} and Table~\ref{tab:ami-trans}) with their corresponding 3.8-bit and 4.0-bit average precision give the lowest WERs. Both the KL and curvature based mixed precision quantized Transformer LMs (LM 17, 19 in Table~\ref{tab:ami-trans}) with 4-bit average precision produced no statistically significant recognition error rate increase relative to the full precision Transformer LM (LM 1 in Table~\ref{tab:ami-trans}) across all test sets.
\end{itemize}

\vspace{-0.5em}
\section{Conclusion}
This paper presents a set of novel mixed precision based neural network LM quantization techniques for LSTM-RNNs and Transformers. In order to account for the locally varying performance sensitivity to low-bit quantization, the optimal local precision settings are automatically learned by either minimizing the KL-divergence or log-likelihood curvature based performance sensitivity measures, or derived using mixed precision neural architecture search.  Quantized LSTM-RNN and Transformer LM parameters are estimated efficiently using alternating direction methods of multipliers based optimization, to address the low convergence speed issue when directly applying gradient descent methods to estimate discrete quantized neural network parameters. Experimental results conducted on two state-of-the-art speech recognition tasks suggest the proposed mixed precision neural network LM quantization methods outperform traditional uniform precision based quantization approaches, and can produce “lossless” quantization and large model size compression ratios of up to around 16 times over the full precision LSTM-RNN and Transformer LM baselines while incurring no statistically significant recognition accuracy degradation. 

All of the three automatic mixed precision configuration approaches proposed in this paper aim to minimize the performance sensitivity to quantization errors. To the best of our knowledge, they are the first research work to apply mixed precision quantization to neural language models for speech recognition tasks. For all the three approaches such performance sensitivity measure is related to the empirical training data log-likelihood degradation given a particular quantization precision bit-width setting relative to the full precision model. The precise manner with which such measure is computed differs among the three. This leads to the measures of information loss via KL-divergence, rate of log-likelihood degradation using curvature, and penalized log-likelihood based mixed precision architecture search approaches respectively. Experimental results obtained both the Switchboard and AMI tasks shown in Tables II to V on LSTM-RNN and Transformer LMs suggest the performance between the KL-divergence and log-likelihood curvature methods is consistently close. This is expected as both are measuring similar forms of log-likelihood based quantization loss, while an extra precision penalty term is introduced in the mixed precision NAS cost function. The larger performance difference between the mixed precision architecture search approach and the other two methods, for example, shown in Tables II and III, may be attributed to the use of Softmax function based architecture weights in the DARTS super-network. When similar architecture weights are obtained, the confusion over different candidate precision settings increases and search errors may occur. 

Future research will focus on improving mixed precision quantization methods and their application to other neural network components of speech recognition systems and end-to-end based neural architectures. 


%



\vspace{-1em}
\section*{Acknowledgment}
This research is supported by Hong Kong Research Grants Council GRF grant 
No. 14200218, 14200220, 14200021, Innovation 
and Technology Fund grant No. ITS/254/19, and Shun Hing Institute
 of Advanced Engineering grant No. MMT-p1-19.

\ifCLASSOPTIONcaptionsoff
  \newpage
\fi



\vspace{-1em}
\nocite{*}
\bibliographystyle{ieeetr}
\bibliography{bare}
%

%








\end{document}